\documentclass[10pt,twocolumn,letterpaper]{article}

\usepackage[pagenumbers]{cvpr} % To force page numbers, e.g. for an arXiv version

\usepackage[utf8]{inputenc} % allow utf-8 input
\usepackage[T1]{fontenc}    % use 8-bit T1 fonts
\usepackage{url}            % simple URL typesetting
\usepackage{amsfonts}       % blackboard math symbols
\usepackage{nicefrac}       % compact symbols for 1/2, etc.
\usepackage{microtype}      % microtypography

\usepackage{tabularx}

\usepackage{natbib}
\usepackage{amsmath,amssymb,amsfonts}
\usepackage{algorithmic}
\usepackage{graphicx}
\usepackage{textcomp}
\usepackage{xcolor}
\usepackage{booktabs}
\usepackage{enumitem}
\usepackage{tabularx} % For width-adjustable tables
\usepackage{booktabs}
\usepackage{makecell} % Add this to the document preamble

\usepackage{pifont} % Add this to your preamble
\usepackage{amssymb}  % For check mark symbols
\usepackage{xcolor}   % For color definitions
\usepackage{amssymb}
\usepackage{xcolor}
\definecolor{mygreen}{RGB}{85, 107, 47} % Darker green
\definecolor{myred}{RGB}{178, 34, 34}   % Darker red
\newcommand{\cmark}{\textcolor{mygreen}{\checkmark}} % Custom green check mark
\newcommand{\xmark}{\textcolor{myred}{\ding{55}}}    % Custom red cross

\usepackage{tabularx}
\usepackage{colortbl}
\usepackage{xcolor}
\usepackage{booktabs}
\usepackage{multirow}

\usepackage{array}

\newcommand{\tableCellHeight}{1}
\newcommand{\tabstyle}[1]{
  \setlength{\tabcolsep}{#1}
  \renewcommand{\arraystretch}{\tableCellHeight}
  \centering
  \small
}

\usepackage{arydshln}

\definecolor{cvprblue}{rgb}{0.21,0.49,0.74}
\usepackage[pagebackref,breaklinks,colorlinks,allcolors=cvprblue]{hyperref}

\def\paperID{2336} % *** Enter the Paper ID here
\def\confName{CVPR}
\def\confYear{2025}

\title{Video-MMMU: Evaluating Knowledge Acquisition \\from Multi-Discipline Professional Videos}

\author{%
  Kairui Hu$^{1}$ \quad Penghao Wu$^{1}$ \quad Fanyi Pu$^{1}$ \quad Wang Xiao$^{1}$ \quad
  Yuanhan Zhang$^{1}$ \quad Xiang Yue$^{2}$ \quad \\ 
  Bo Li$^{1}$ \quad Ziwei Liu$^{1}$\thanks{Corresponding author}\\
  $^{1}$S-Lab, Nanyang Technological University \\
  $^{2}$Carnegie Mellon University \\
\text{\url{https://videommmu.github.io/}}
}

\usepackage{cuted}
\usepackage{capt-of}

\begin{document}

\twocolumn[{%
   \renewcommand\twocolumn[1][]{#1}%
   \maketitle
   \vspace{-15pt}
   \begin{center}
    \centering
    \includegraphics[width=0.96\textwidth]{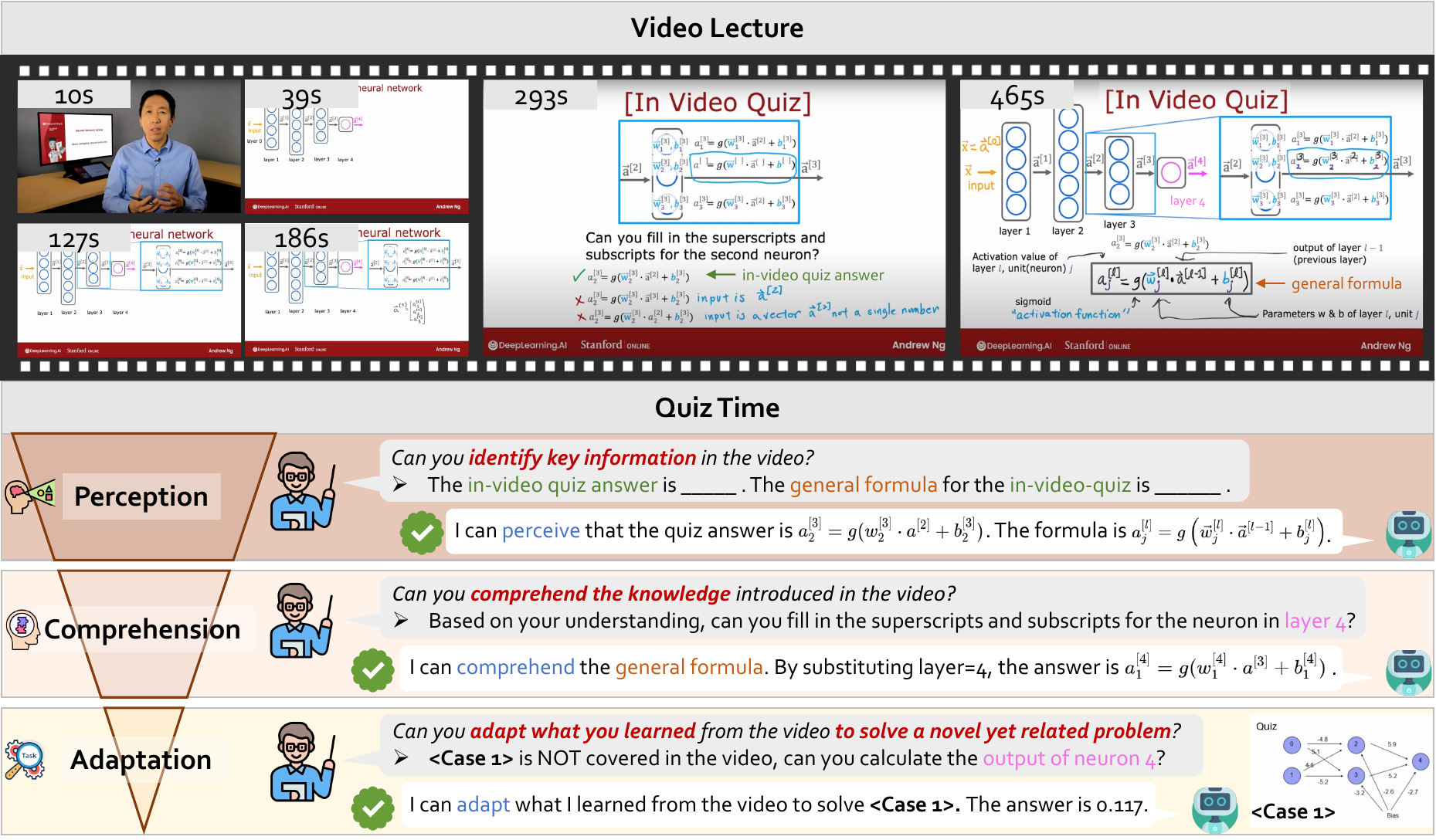}
    \captionof{figure}{An illustration of \textbf{Video-MMMU}: Evaluating the knowledge acquisition capability from videos through three cognitive stages: \textbf{1) Perception: }  if models can identify key information related to knowledge; \textbf{2) Comprehension: }  if models can interpret the underlying concepts; \textbf{3) Adaptation:}  if models can adapt the knowledge from videos to novel scenarios.}
    \label{fig:figure1}
   \end{center}%
  }]

\maketitle

\begin{abstract}
\label{sec:abstract}

Humans acquire knowledge through three cognitive stages: perceiving information, comprehending knowledge, and adapting knowledge to solve novel problems. Videos serve as an effective medium for this learning process, facilitating a progression through these cognitive stages. However, existing video benchmarks fail to systematically evaluate the \textit{knowledge acquisition} capabilities in Large Multimodal Models (LMMs). To address this gap, we introduce Video-MMMU, a multi-modal, multi-disciplinary benchmark designed to assess LMMs' ability to acquire and utilize knowledge from videos. Video-MMMU features a curated collection of 300 expert-level videos and 900 human-annotated questions across six disciplines, evaluating \textit{knowledge acquisition} through stage-aligned question-answer pairs: Perception, Comprehension, and Adaptation. A proposed knowledge gain metric, $\Delta_{\text
{knowledge}}$, quantifies improvement in performance after video viewing. Evaluation of LMMs reveals a steep decline in performance as cognitive demands increase and highlights a significant gap between human and model knowledge acquisition, underscoring the need for methods to enhance LMMs' capability to learn and adapt from videos.

\end{abstract}
\section{Introduction}
\label{sec:introduction}

Humans acquire knowledge through three fundamental cognitive stages outlined in Bloom's taxonomy \cite{forehand2010bloom}: \textbf{1)} perceiving information, \textbf{2)} comprehending knowledge, and \textbf{3)} adapting knowledge to solve novel problems.
Video serves as an ideal medium for this learning process, enabling a natural progression from information intake to practical application, making video-based learning a valuable tool for knowledge acquisition
\cite{videobasedlearning1,videobasedlearning2,videobasedlearning3}. Consider learning neural network forward propagation through video lectures (Fig.~\ref{fig:figure1}): learners first recognize fundamental concepts like activation functions, then demonstrate understanding through exercises, and ultimately apply this knowledge to solve novel exam problems. This progression naturally aligns with Bloom's cognitive stages, providing a systematic framework for assessing knowledge acquisition from videos.
For Large Multimodal Models (LMMs) to operate effectively in the wild like humans, learning from videos is an essential capability for continuous knowledge acquisition. However, existing video benchmarks lack systematic evaluation of this critical ability.
% Despite the importance of video-based learning, current video benchmarks lack systematic evaluation of knowledge acquisition. 
To bridge this gap, we introduce \textbf{Video-MMMU}, a massive multi-modal, multi-disciplinary video benchmark that evaluates the knowledge acquisition capability from educational videos through three main features:
\textbf{1) Knowledge-intensive Video Collection:} Our dataset comprises 300 expert-level videos spanning 6 professional disciplines: Art, Business, Science, Medicine, Humanities, and Engineering, with 30 subjects distributed among them.
\textbf{2) Knowledge Acquisition-based Question Design:} Each video includes three question-answer pairs aligned with the three knowledge acquisition stages: Perception (identifying key information related to the knowledge), Comprehension (understanding the underlying concepts), and Adaptation (applying knowledge to new scenarios).
\textbf{3) Quantitative Knowledge Acquisition Assessment:} We propose a knowledge acquisition metric, denoted as \(\Delta_{\text{knowledge}}\), to measure performance gains on practice exam questions after learning from videos. This metric enables us to quantitatively evaluate how effectively large multimodal models (LMMs) can assimilate and utilize the information presented in the videos to solve real-world, novel problems.

% quantifies how effectively LMMs assimilate and utilize the knowledge from videos.

We evaluate both open-source and proprietary LMMs on Video-MMMU, revealing several key findings:
\textbf{1) Progressive Performance Decline:} Model performance decreases as cognitive demands increase. While models perform relatively better on perception tasks, their accuracy drops notably on comprehension tasks and declines further on adaptation tasks.
\textbf{2) Knowledge Acquisition from videos is Challenging:} The knowledge acquisition metric $\Delta_{\text{knowledge}}$ reveals a significant gap between human and model performance. While humans achieve substantial improvement ($\Delta_{\text{knowledge}} = 33.1\%$) after watching the videos, even the top performing models show smaller knowledge gains (GPT-4o \cite{openai2024gpt4o}: $\Delta_{\text{knowledge}}=15.6\%$, Claude-3.5-Sonnet \cite{sonnet}: $\Delta_{\text{knowledge}}=11.4\%$).
This limitation underscores a challenge in current LMMs. While humans naturally acquire knowledge through video-based learning, having developed this capability through classroom learning and educational experiences throughout life, LMMs struggle to effectively learn from videos. These findings emphasize the need for further research to enhance how LMMs acquire and utilize video-based information, bringing them closer to human-level learning processes.

\begin{figure*}[ht]
    \centering
    \includegraphics[width=\textwidth]{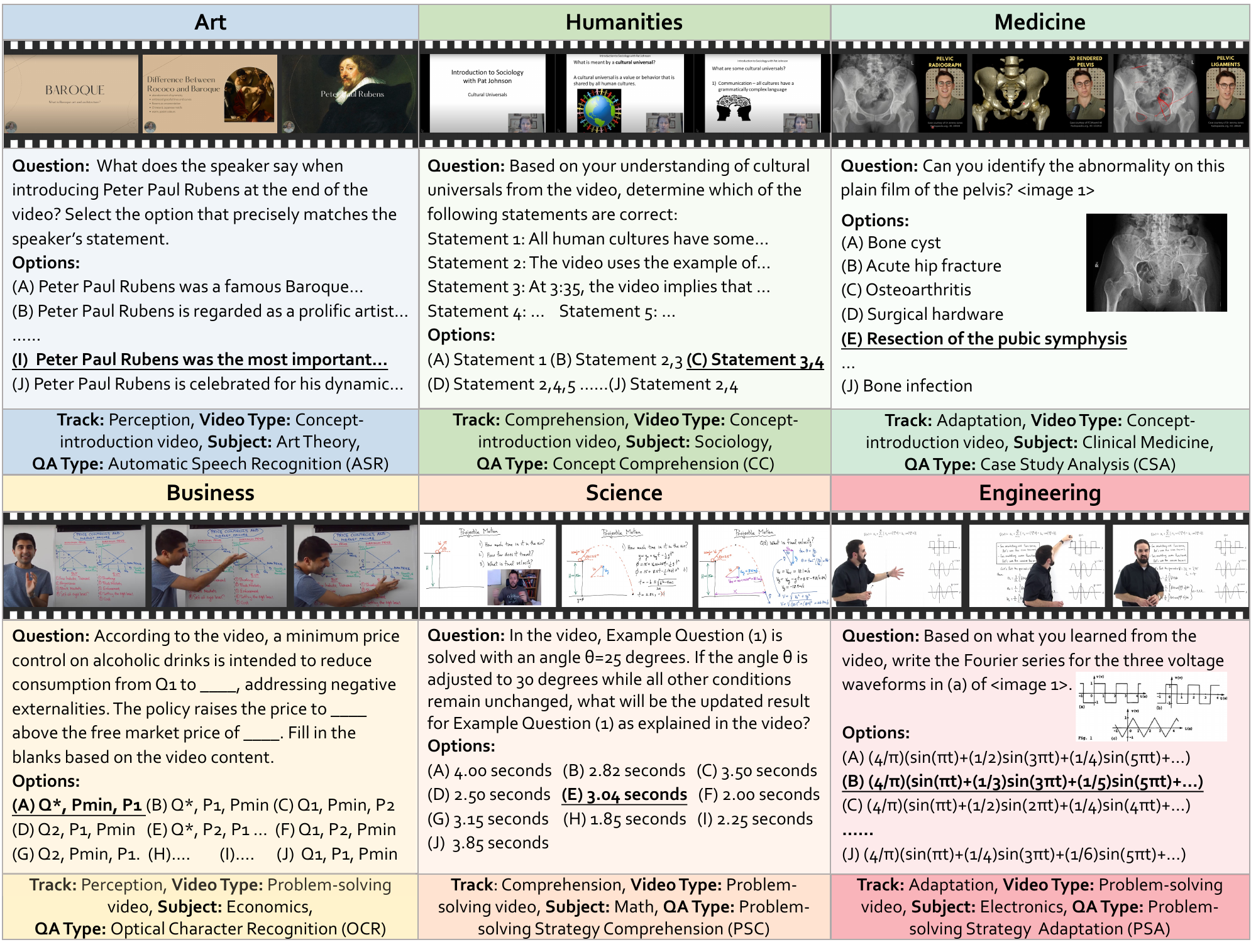}
    \vspace{-20pt}
    \caption{Sampled Video-MMMU examples across 6 academic disciplines and 3 tracks. The examples are organized in two rows based on distinct video types: (1) Concept-Introduction videos (top row) focus on teaching factual knowledge, fundamental concepts, and theories through explanatory content, while (2) Problem-Solving videos (bottom row) demonstrate step-by-step solutions to an example question.}
    \label{fig:figure2}
\end{figure*}

\section{Related Work}
\label{sec:related}

\subsection{VideoQA Benchmarks}
\label{sec:videoqa benchmark}
Existing video benchmarks focus primarily on visual understanding tasks, including action understanding \cite{yu2019activityqa, patraucean2023perception, egoschema, xiao2021next, funqa, Khattak2024cvrres}, temporal reasoning \cite{liu2024tempcompass, li2024vitatecsdiagnosticdatasettemporal, wang2024lvbench, wu2024longvideobench, cai2024temporalbench, clever, song2023moviechat}, and video captioning \cite{xu2017video, Xu_2016_CVPR, vatex, youcook2, chai2024auroracap}.
Several benchmarks enhance scene interpretation by incorporating external knowledge, including KnowIT-VQA \cite{garcia2020knowit} and WorldQA \cite{zhang2024WorldQA}. Recent benchmarks like Video-MME \cite{fu2024video}, MMBench-Video \cite{fang2024mmbenchvideo}, and MLVU \cite{MLVU} have expanded the scope to assess multi-tasking and multi-domain video understanding. While these benchmarks recognize videos as visual scenes for interpretation, Video-MMMU uniquely recognizes video as an educational medium, emphasizing knowledge-driven question-answering on videos.

\subsection{Knowledge-driven Benchmarks}
As AI systems progress toward Expert AGI \cite{pmlr-v235-morris24b}, knowledge-driven benchmarks have emerged to evaluate models' professional expertise. Early benchmarks such as AGIEval \cite{zhong-etal-2024-agieval} and ARC \cite{arc} focus on standardized exams and science questions, respectively. MMLU \cite{hendryckstest2021} expands evaluation across STEM disciplines, while MMLU-Pro \cite{wang2024mmlu} introduces more challenging reasoning-focused questions.
Multi-modal benchmarks extend this evaluation scope further. ScienceQA \cite{scienceqa} assesses multi-modal reasoning on elementary to high-school science questions. MMMU \cite{yue2023mmmu} advances to college-level questions requiring subject-specific knowledge and deliberate reasoning. MMMU-Pro \cite{yue2024mmmu} enhances MMMU questions for more robust evaluation. While these benchmarks evaluate models' pre-trained knowledge and reasoning abilities on text and images, Video-MMMU uniquely focuses on assessing how effectively models can acquire and apply knowledge from videos.
\section{Video-MMMU Dataset}
\label{sec:method}

\begin{figure*}
    \centering
    \includegraphics[width=\textwidth]{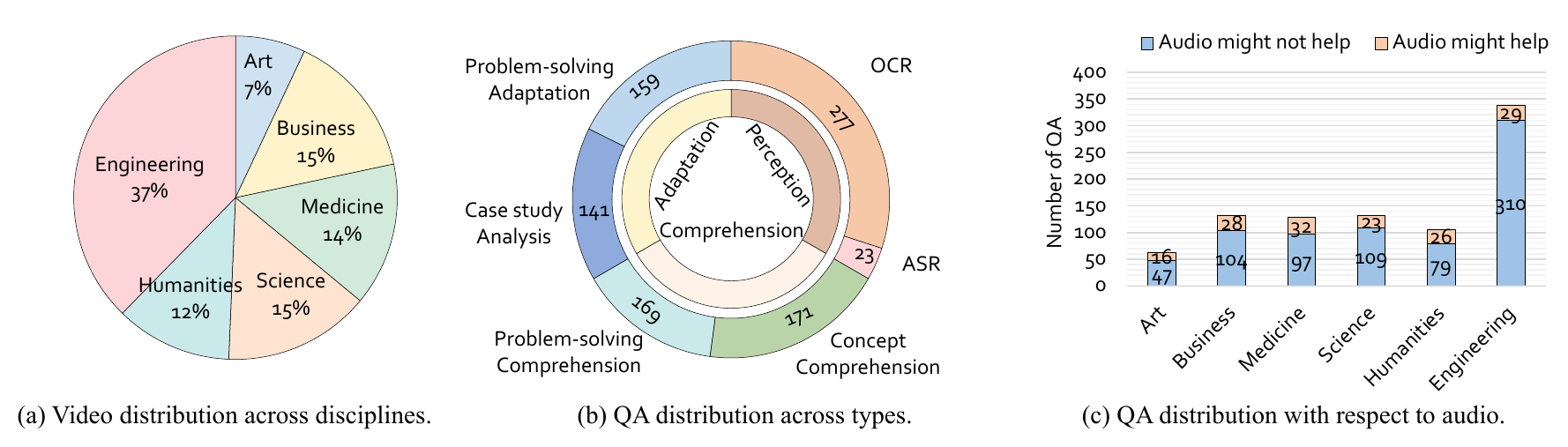}
    \caption{Taxonomy of QA types and video disciplines.}
    \label{fig:figure3}
\end{figure*}

We introduce Video-MMMU (Massive Multi-discipline Multimodal Understanding), a video benchmark designed to evaluate knowledge acquisition from educational videos across 30 subjects in 6 professional disciplines: Art, Business, Medicine, Science, Humanities, and Engineering. The video distribution across disciplines is shown in Fig. \ref{fig:figure3}\textcolor{cvprblue}{a}.

\subsection{Video Collection}
\label{subsec:videocollection}
The dataset consists of 300 college-level educational videos, systematically curated through a rigorous three-phase process:
\textbf{1) Topic Selection:} Domain experts conduct a comprehensive analysis of college curricula across 30 subjects, establishing a diverse pool of 450 foundational assessment topics.
\textbf{2) Video Curation:} Leveraging GPT-4o \cite{openai2024gpt4o}, we generated 10 search queries per topic. These search queries are processed through the YouTube Data API to create an initial candidate video pool.
\textbf{3) Quality Assurance:} We implemented a three-tier review protocol:
First, annotators cross-check to filter out videos with poor audio-visual quality or irrelevant content. Second, we employ GPT-4o \cite{openai2024gpt4o} to assess the technical depth of the videos by analyzing 10 sampled frames from each video. We prioritize in-depth lectures, tutorials, and detailed problem-solving demonstrations while excluding beginner-level introductions and superficial overviews. Finally, domain experts verify alignment with college curriculum standards and confirm appropriate domain knowledge depth.

The Video-MMMU dataset comprises two distinct categories:
\textbf{1) Concept-introduction Videos:} These videos provide comprehensive explanations of factual knowledge, including fundamental concepts and theories.
\textbf{2) Problem-solving Videos:} These videos demonstrate step-by-step problem solutions, particularly in STEM disciplines where systematic reasoning and detailed calculations are required.

\subsection{QA Annotation}

\subsubsection{QA Taxonomy}
We annotate questions across three cognitive stages: Perception, Comprehension, and Adaptation, each assessing progressively deeper levels of knowledge acquisition.

\noindent \textbf{Perception Questions} assess the ability to perceive information from videos through:
\textbf{1) Optical Character Recognition (OCR):} These questions require identifying and extracting key details from visual content, including formulas, data points, charts, and handwritten notes. An example is shown in Fig.~\ref{fig:figure2} (Business), where the question requires extracting multiple economic variables from handwritten notes.
\textbf{2) Automatic Speech Recognition (ASR):} These questions assess the ability to accurately transcribe spoken content into text, as illustrated in Fig.~\ref{fig:figure2} (Art).

\noindent \textbf{Comprehension Questions} evaluate the ability to understand knowledge presented in videos through:
\textbf{1) Concept Comprehension (CC):} These questions assess understanding of concepts introduced in the videos. We primarily use a multiple-answer multiple-choice (MAMC) format, where each question presents 4-10 statements about video content, with multiple correct statements possible. As shown in Fig.~\ref{fig:figure2} (Humanities), one must identify all correct statements about the video content to demonstrate a comprehensive understanding.
\textbf{2) Problem-solving Strategy Comprehension (PSC):} For videos demonstrating step-by-step solutions to example questions, an intuitive way to assess the understanding of the solution is to test the same question with different input values. As illustrated in Fig.~\ref{fig:figure2} (Science), when a video demonstrates trajectory time calculation with a 25-degree angle, the question changes this to 30 degrees. This approach verifies comprehension of the underlying solution strategy rather than the memorization of answers. The cognitive difficulty lies between perception and adaptation, requiring new calculations while following the same reasoning process in the video.

\noindent \textbf{Adaptation Questions} assess the ability to adapt video knowledge to new scenarios:
\textbf{1) Case Study Analysis (CSA):} These questions evaluate the application of concepts to novel real-world scenarios. As shown in Fig.~\ref{fig:figure2} (Medicine), while the video explains various pelvic pathologies, the question requires analysis of a new patient's pelvic radiograph to identify specific abnormalities. This tests the model's ability to adapt theoretical knowledge from videos to practical clinical diagnosis.
\textbf{2) Problem-solving Strategy Adaptation (PSA):} These questions evaluate how learners adapt learned solution methods to new problems. For instance, in Fig.~\ref{fig:figure2} (Engineering), the video demonstrates the calculation of Fourier series for one type of waveform, while the question presents a different waveform pattern. To solve this new problem, one needs to identify key similarities and differences between the video example and the new problem, then adjust the solution method accordingly. The distribution of these question types is illustrated in Fig.~\ref{fig:figure3}\textcolor{cvprblue}{b}.

\subsubsection{Annotation and Quality Control}
\noindent \textbf{Annotation Process:} Our annotation follows a multi-stage process to ensure quality: \textbf{1) Initial Annotation:} Annotators thoroughly review each video and annotate three questions aligned with our cognitive tracks, following the QA taxonomy shown in Fig.\ref{fig:figure3}\textcolor{cvprblue}{b}. To enhance assessment rigor, we annotate 10 options for each multiple-choice question (MCQ). \textbf{2) Quality Assurance:} Firstly, annotators cross-check each other's questions for consistency and clarity. Secondly, QA pairs are processed by OpenAI o1\cite{o1} to refine the language and verify the correctness of ground-truth answers. Thirdly, domain experts review each question for technical accuracy and alignment with the intended cognitive stages. For Adaptation questions, experts verify that the question tests the same knowledge presented in the video but in a novel scenario, ensuring they utilize the same concepts, formulas, or similar problem-solving strategies. Finally, we employ Gemini 1.5 Pro~\cite{gemini2023family} to analyze each video-question pair and determine whether audio might be helpful to solve the question, as shown in Fig.~\ref{fig:figure3}\textcolor{cvprblue}{c}. This analysis will benefit more future Large Multimodal Models (LMMs) with audio processing capabilities.

\noindent \textbf{Question Sources:} For the Perception and Comprehension tracks, questions are manually created by our annotators. For the Adaptation track, which requires practical problems from exams and case studies, our approach varies by discipline. In Science, Engineering, Medicine, and Business, we source questions from MMMU~\cite{yue2023mmmu} and MMMU-pro~\cite{yue2024mmmu}, which provide validated college exam questions well suited for testing knowledge adaptation. For Art and Humanities, where adaptation requires more context-dependent assessment, we manually create case study questions to ensure alignment with video concepts.

\subsection{Comparison with Existing Benchmarks}
\label{subsec:compare}
Video-MMMU distinguishes itself through its emphasis on how models can learn and apply knowledge from professional educational videos. Our videos feature comprehensive lectures, tutorials, and step-by-step problem-solving demonstrations, delivering dense information through multiple visual formats, including charts, diagrams, and handwritten explanations. With an average duration of 506.2 seconds, the videos provide extensive coverage of domain-specific knowledge across various disciplines.
As shown in Table~\ref{tab:video_statistics}, our questions are substantially longer than existing benchmarks, averaging 75.7 words per question, reflecting the complexity of knowledge-driven evaluation. We systematically evaluate knowledge acquisition from videos through three cognitive stages. The Adaptation track advances video-based learning evaluation beyond basic content understanding to assess how effectively models can apply the acquired knowledge to novel problems.

\begin{table}[t]
\tabstyle{2pt}
\centering
\resizebox{0.96\linewidth}{!}{%
\begin{tabular}{@{}l c c c c@{}}
\toprule
\textbf{Benchmarks} & \textbf{Video} & \textbf{Question}  & \textbf{Video} & \textbf{Knowledge} \\
 & \textbf{Domain} & \textbf{Length} & \textbf{Duration}  & \textbf{driven}\\
\midrule

Video-MME \cite{fu2024video}  & Open & 35.7     & 1017.9 & \xmark \\
MMBench-Video \cite{fang2024mmbenchvideo}  & Open & 10.9   & 165.4 & \xmark \\
Video-Bench \cite{videobench} & Open & 21.3 & 56.0 & \xmark \\
TempCompass \cite{liu2024tempcompass}       & Open  & 49.2 & 11.4& \xmark \\
MVBench \cite{Li_2024_CVPR}        & Open & 27.3    & 16.0 & \xmark \\
AutoEval-Video \cite{autoevalvideo} & Open  & 11.9 & 14.6 & \xmark \\
\midrule
Video-MMMU  & Professional & 75.7  & 506.2 & \cmark  \\

\bottomrule
\end{tabular}
}
\caption{Comparison of Video-MMMU and other widely adopted video benchmarks.}
\label{tab:video_statistics}
\end{table}

\section{Experiments}
\label{sec:exp}

\begin{table*}[t]
\renewcommand{\arraystretch}{1.2}
%\tabstyle{13pt}
\small
\centering
\begin{tabular}{@{}l @{\hspace{1pt}}c c @{\hspace{3pt}}c @{\hspace{3pt}}c !{\vrule width 0.5pt} c c c c c c@{}}
\toprule
\textbf{Model} & \textbf{Overall} & \multicolumn{3}{c}{\textbf{Results by Track}} & \multicolumn{6}{c}{\textbf{ Results by Discipline}} \\
\cmidrule(lr){3-5} \cmidrule(lr){6-11}
&&\textbf{Perception} & \textbf{Comprehension} & \textbf{Adaptation}& \textbf{Art.} & \textbf{Biz.} & \textbf{Sci.}& \textbf{Med.} & \textbf{Hum.} & \textbf{Eng.} \\
\midrule
\textbf{Random Choice} & 14.00 & 12.00 & 14.00 & 16.00 & 11.11 & 12.88 & 12.12 & 22.48 & 10.48 & 13.57 \\
\textbf{Human Expert} & 74.44 & 84.33 & 78.67 & 60.33 & 80.95 & 78.79 & 74.24 & 70.54 & 84.76 & 69.91 \\

\midrule
\rowcolor{blue!20} \multicolumn{11}{l}{\textbf{Proprietary LMMs}} \\
\midrule
\textbf{Gemini 1.5 Flash \cite{gemini2023family}}  & 49.78 & 57.33 & 49.00 & 43.00 & 63.49 & 53.03 & 43.18 & 49.61 & 59.05 & 45.72 \\
\textbf{Gemini 1.5 Pro \cite{gemini2023family}}  & 53.89 & 59.00 & 53.33 & 49.33 & 57.14 & 59.09 & 49.10 & 57.42 & 58.10 & 50.31 \\

% \textbf{Claude-3-Opus} & -- & 1 fps & 50.44 & 48.15 & 58.73 & 44.44 \\

% \textbf{GPT4o-mini} & -- & 50 & 53.09 & 60.85 & 53.97 & 44.44 \\
\textbf{GPT-4o \cite{openai2024gpt4o}} & 61.22 & 66.00  & 62.00 & \textbf{55.67} & 69.52 & 66.88 & 51.55 & \textbf{64.76} & 69.52 & 57.13 \\
\textbf{Claude-3.5-Sonnet \cite{sonnet}} & \textbf{65.78} & \textbf{72.00} & \textbf{69.67} & \textbf{55.67} & 66.67 & \textbf{75.00} & \textbf{56.06} & 58.14 & \textbf{75.24} & \textbf{66.08} \\

\midrule
\rowcolor{blue!10} \multicolumn{11}{l}{\textbf{Open-source LMMs}} \\
\midrule
% \textbf{LLaVA-OneVision} & 7B & 8 & 47.26 & 58.20 & 42.86 & 40.74 \\
\textbf{VILA1.5-8B \cite{lin2023vila}}  & 20.89 & 20.33 & 17.33 & 25.00 & 34.92 & 14.39 & 19.70 & 19.38 & 21.91 & 21.53 \\
\textbf{LongVA-7B \cite{zhang2024longva}} & 23.98 & 24.00 & 24.33 & 23.67 & 41.27 & 20.46 & 21.97 & 24.03 & 23.81 & 23.01 \\

\textbf{Llama-3.2-11B \cite{meta2024llama32}} & 30.00 & 35.67 & 32.33 & 22.00 & 39.68 & 28.79 & 21.21 & 35.66 & 33.33 & 28.91 \\

\textbf{LLaVA-OneVision-7B \cite{li2024llava}} & 33.89 & 40.00 & 31.00 & 30.67 & 49.21 & 29.55 & 34.85 & 31.78 & 46.67 & 29.20 \\
\textbf{VILA1.5-40B \cite{lin2023vila}} & 34.00 & 38.67 & 30.67 & 32.67 & 57.14 & 27.27 & 23.49 & 37.99 & 41.91 & 32.45 \\
\textbf{LLaVA-Video-7B \cite{zhang2024videoinstructiontuningsynthetic}} & 36.11 & 41.67 & 33.33 & 33.33 & 65.08 & 34.09 & 32.58 & 42.64 & 45.71 & 27.43 \\
\textbf{InternVL2-8B \cite{chen2024far}} & 37.44 & 47.33 & 33.33 & 31.67 & 55.56 & 34.09 & 30.30 & 34.11 & 41.91 & 38.05 \\
\textbf{MAmmoTH-VL-8B \cite{guo2024mammothvlelicitingmultimodalreasoning}} & 41.78 & 51.67 & 40.00 & 33.67 & 47.62 & 37.88 & 36.36 & 36.43 & 49.52 & 43.95 \\

\textbf{LLaVA-OneVision-72B \cite{li2024llava}}  & 48.33 & 59.67 & 42.33 & 43.00 & 61.91 & 46.21 & 40.15 & 54.26 & 60.00 & 43.95\\
\textbf{LLaVA-Video-72B \cite{zhang2024videoinstructiontuningsynthetic}} & 49.67 & 59.67 & 46.00 & 43.33 & 69.84 & 44.70 & 41.67 & 58.92 & 57.14 & 45.13 \\
\textbf{Aria \cite{aria}} & 50.78 & 65.67 & 46.67 & 40.00 & \textbf{71.43} & 47.73 & 44.70 & 58.92 & 62.86 & 43.66  \\

\bottomrule
\end{tabular}
\caption{Video-MMMU Evaluation Results across three cognitive tracks (Perception, Comprehension, Adaptation) and six disciplines (Art, Business, Science, Medicine, Humanities, Engineering).}
\label{tab:trackresult}
\end{table*}

\subsection{Settings}

\noindent \textbf{Baselines. } 
We evaluate open-source LMMs including LLaVA-OneVision~\cite{li2024llava}, LLaVA-Video~\cite{zhang2024videoinstructiontuningsynthetic}, LongVA~\cite{zhang2024longva}, VILA-1.5~\cite{lin2023vila}, Qwen2-VL~\cite{Qwen2VL}, InternVL2~\cite{chen2024far}, Llama-3.2~\cite{meta2024llama32}, MAmmoTH-VL \cite{guo2024mammothvlelicitingmultimodalreasoning}, Aria~\cite{aria}; and proprietary models GPT-4o~\cite{openai2024gpt4o}, Gemini 1.5 Pro~\cite{gemini2023family}, Gemini 1.5 Flash~\cite{gemini2023family}, Claude-3.5-Sonnet~\cite{sonnet}. The numbers of sampled frames are 32 for LLaVA-OneVision, 64 for LLaVA-Video, 64 for LongVA, 32 for VILA-1.5, 32 for InternVL2, 10 for Llama-3.2, 32 for MAmmoTH-VL, 64 for Aria, 20 for Claude-3.5-Sonnet, 50 for GPT-4o.

\noindent \textbf{Human Experts. } 
To assess the performance of Human Experts, we recruited senior undergraduate students and instructed them to complete the following tests: The students first attempted the Adaptation question without viewing the videos. Subsequently, they watched each assigned video and answered the corresponding Perception, Comprehension, and Adaptation questions. While students could refer to course materials and notes, they were not allowed to search for answers on the Internet.

\noindent \textbf{Inputs. } 
We provide videos and questions as inputs for the Perception and Comprehension tracks. For the Adaptation track, we append the question's image to the end of each video. We add a prompt to indicate that the image for the Adaptation track question appears in the final frame.
% In the future, if LMMs can handle image and video inputs concurrently, this process will no longer be necessary.
% We follow the format: [Video] [Image (Only for Adaptation)] [Question] [Options]. 
% We introduce an additional pre-prompt for the Adaptation track:
% \textit{You should watch and learn from the video. Then, adapt what you learned to answer the following multi-choice question. The image for this question is at the end of the video. [Question][Options]}.

\noindent \textbf{Evaluations. }
We evaluate model outputs using an automated, rule-based pipeline. The system employs regular expressions to extract key elements such as option letters and numerical values. Responses lacking valid answers are marked as incorrect. 
We use the micro-averaged accuracy as our evaluation metric. The evaluation is conducted using LMMs-Eval \cite{lmmseval}.

\subsection{Main Results}
\subsubsection{Performance by Track}

\noindent \textbf{Human vs. Model Performance:} Human experts outperform models across all tracks, with Claude achieving the highest model scores but still showing a gap to humans. Both humans and models exhibit declining performance from perception through comprehension to adaptation, indicating that deeper cognitive stages require more advanced capabilities.
\noindent \textbf{Perception Track:} Many models achieve an accuracy over 50\%, suggesting perception is a more fundamental capability among the three stages.
\noindent \textbf{Comprehension Track:} Comprehending college-level knowledge from videos requires pre-trained knowledge as a foundation. Compared to the Perception score, most open-source models show a $10\sim 20\%$ decline in Comprehension score, while proprietary models show less performance decline and generally achieve higher comprehension scores, demonstrating their superior capabilities in comprehending knowledge-intensive videos.
\noindent \textbf{Adaptation Track:} Adaptation emerges as the most challenging stage, with most models scoring below 50\%. Even top-performing models like Claude-3.5-Sonnet exhibit a substantial performance decline in Adaptation. This indicates a natural gap between theoretical understanding and practical application. While models might understand the knowledge from videos at a surface level, they currently lack the advanced capability to effectively acquire and apply what they learned from the video to solve practical problems.

\subsubsection{Performance by Discipline}
Model performance varies across disciplines. 
Models demonstrate superior performance in Art and Humanities disciplines, where videos primarily focus on conceptual presentation. In comparison, they achieve lower accuracies in Science, Engineering, Business, and Medicine, which demand quantitative reasoning and interpretation of detailed technical visuals such as diagrams and handwritten notes. This performance differential suggests models are generally more adept at processing factual knowledge but underperform in domains requiring complex computation, deliberate reasoning, and visual analysis.

\begin{figure}
    \centering
    \includegraphics[width=0.47\textwidth]{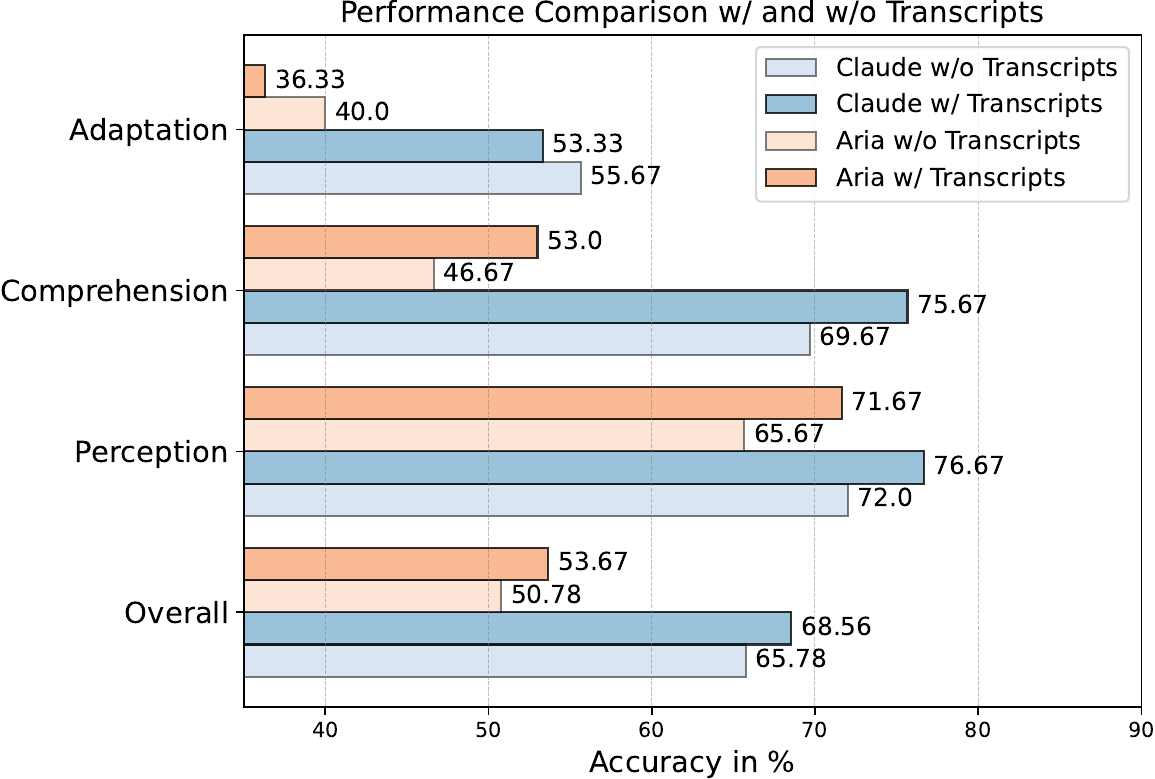}
    \caption{Performance comparison across tracks before and after adding audio transcripts.}
    \label{fig:audio}
\end{figure}

\begin{figure*}[t]
\begin{subfigure}[b]{0.48\textwidth}
    \centering
    \includegraphics[width=\textwidth]{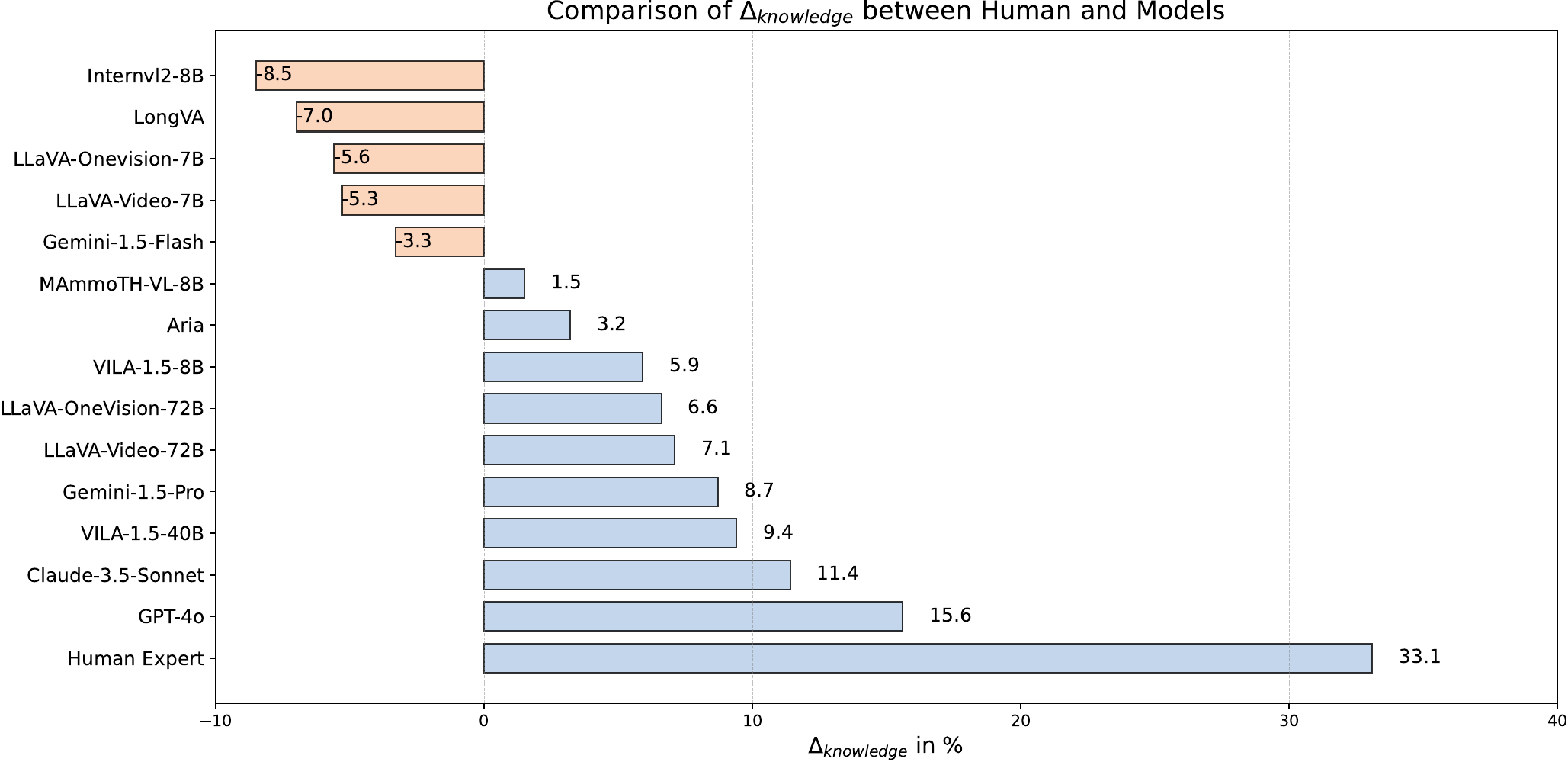}
    \caption{Comparison of $\Delta_{\text{knowledge}}$ (performance improvement in the Adaptation track after watching the video compared to before).}
    \label{fig:figure4}
\end{subfigure}
\hfill
\begin{subfigure}[b]{0.48\textwidth}
    \centering
    \includegraphics[width=\textwidth]{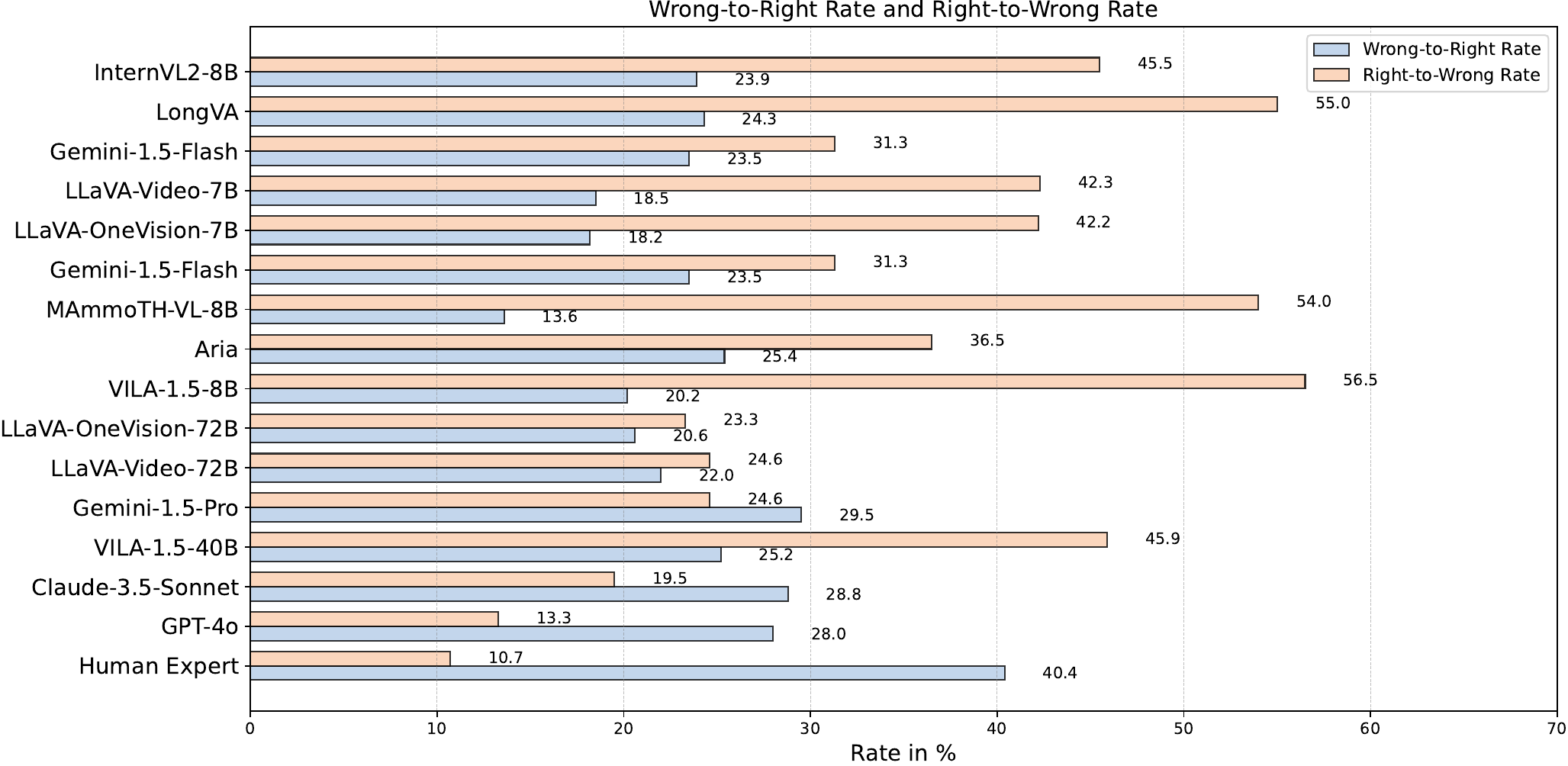}
    \caption{Comparison of Wrong-to-Right Rate (the percentage of Adaptation track questions that were initially answered incorrectly without the video but correctly after watching the video) and Right-to-Wrong Rate (vice versa).}
    \label{fig:figure5}
    
\end{subfigure}

\vspace{-2pt}
\caption{Key findings in the experiment of $\Delta_{\text{knowledge}}$.}
\label{fig:furtheranalysis}
\vspace{-12pt}
\end{figure*}

\subsection{Impact of Audio Transcript} Audio conveys information in knowledge-intensive videos. To study the impact of audio transcripts, we use OpenAI Whisper \cite{whisper} to generate audio transcripts and append them to the input prompt. We conduct evaluation on the top-performing open-source model Aria \cite{aria} and proprietary model Claude-3.5-Sonnet \cite{sonnet}.

As shown in Fig.~\ref{fig:audio}, audio transcripts yield overall performance improvements across different evaluation tracks. In the Comprehension track, the enhancement is most pronounced, reflecting audio's contribution to video content understanding. Similarly, the Perception track demonstrates performance gains, suggesting that audio enhances information extraction from videos. The Adaptation track, however, shows a decrease in performance. This decline indicates that while audio enriches basic understanding, it might complicate the adaptation of knowledge to novel scenarios. These contrasting effects reveal a trade-off: audio transcripts enhance immediate comprehension but potentially constrain models' ability to adapt knowledge to new scenarios.

\section{Knowledge Acquisition in Adaptation Track}

\subsection{Settings}
We introduce a knowledge acquisition metric $\Delta_{\text{knowledge}}$ to measure how much knowledge models gain from videos through their performance improvement on practical exam questions in the Adaptation track. We define $\Delta_{\text{knowledge}}$ as:

\[
\Delta_{\text{knowledge}} = \frac{\text{Acc}_{\text{post}} - \text{Acc}_{\text{pre}}}{100\% - \text{Acc}_{\text{pre}}} \times 100\%
\]
where $\text{Acc}_{\text{pre}}$ and $\text{Acc}_{\text{post}}$ represent the accuracy before and after watching the video, respectively. This normalized metric accounts for different baseline difficulty levels. For example, improving from 90\% to 95\% ($\Delta_{\text{knowledge}} = 50\%$) indicates more substantial video-based learning than improving from 0\% to 5\% ($\Delta_{\text{knowledge}} = 5\%$). We evaluate $\Delta_{\text{knowledge}}$ across top-performing open-source and proprietary models.

\subsection{Findings}
\noindent \textbf{Human-Model Knowledge Acquisition Gap:} Fig.~\ref{fig:figure4} reveals a substantial disparity between human and model learning capabilities. Humans demonstrate a $\Delta_{\text{knowledge}}$ of 33.1\% after viewing the videos, while the best-performing model GPT-4o achieves only 15.6\%. Some models even exhibit negative $\Delta_{\text{knowledge}}$, suggesting their performance declines after video exposure.

This gap highlights a fundamental challenge in current models. Humans naturally acquire information through video-based learning, having developed this capability through classroom education and video content throughout their lives. While many models can process video information, they struggle to effectively learn new knowledge from the video and apply it in practice.

\begin{figure*}
    \centering
    \includegraphics[width=\textwidth]{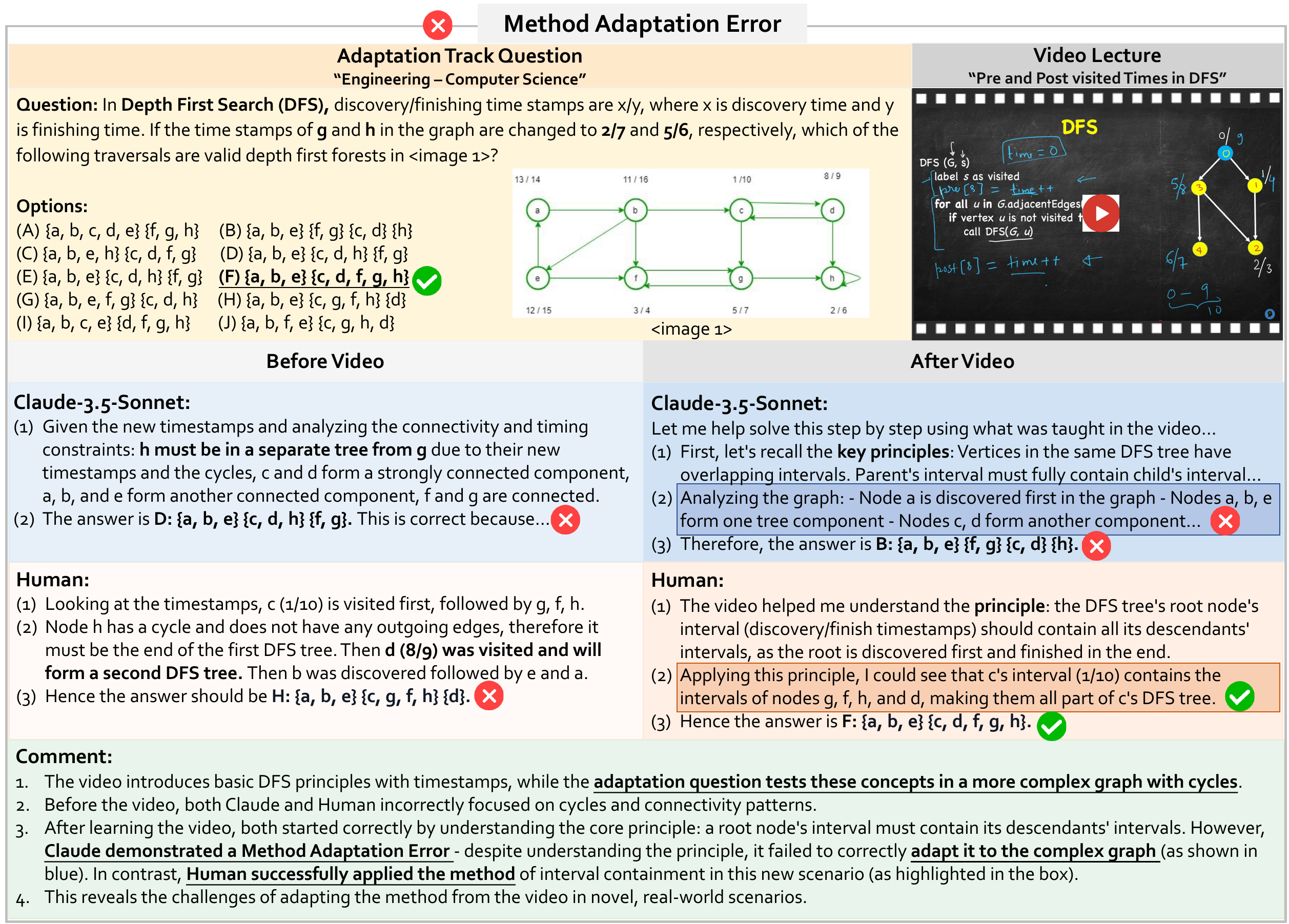}
    \caption{A Case of Method Adaptation Error. The model can recall the correct knowledge from the video but fails to adapt the method to a new scenario. More error cases are analyzed in the Appendix.}
    \label{fig:case_study}
\end{figure*}

\begin{figure}
    \centering
    \includegraphics[width=0.47\textwidth]{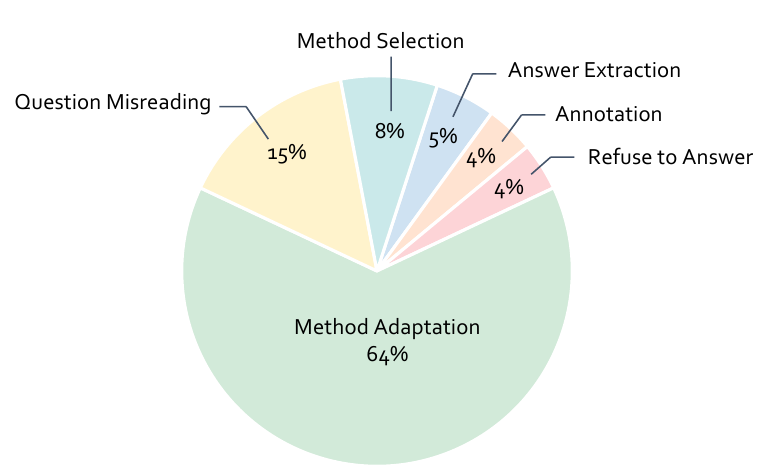}
    \caption{Distribution of the 100 human-annotated errors in Claude-3.5-Sonnet.}
    \label{fig:errorpiechart}
\end{figure}

\noindent \textbf{Video Impact on Model Responses:} 
While low $\Delta_{\text{knowledge}}$ scores might suggest limited net knowledge gain, models' responses change substantially after watching the videos. As shown in Fig.~\ref{fig:figure5}, we analyze these changes through two metrics: Wrong-to-Right Rate (the percentage of questions initially answered incorrectly but correctly after watching videos) and Right-to-Wrong Rate (the percentage of questions correctly answered before but incorrectly after watching videos). We define the Wrong-to-Right Rate as:

\[
\text{Wrong-to-Right Rate} = \frac{N_{\text{Wrong-to-Right}}}{N_{\text{Wrong-before}}} \times 100\%
\]
, where $N_{\text{Wrong-to-Right}}$ refers to the number of questions that were answered incorrectly before watching the video but correctly after watching the video, and $N_{\text{Wrong-before}}$ is the total number of questions that were answered incorrectly before watching the video. 

Similarly, we define the Right-to-Wrong Rate as:

\[
\text{Right-to-Wrong Rate} = \frac{N_{\text{Right-to-Wrong}}}{N_{\text{Right-before}}} \times 100\%
\]
, where $N_{\text{Right-to-Wrong}}$ refers to the number of questions that were answered correctly before watching the video but incorrectly after watching the video, and $N_{\text{Right-before}}$ is the total number of questions that were answered correctly before watching the video.

Interestingly, models achieve moderate Wrong-to-Right Rates (e.g., Gemini-1.5-Pro: 29.5\%), demonstrating certain ability to acquire knowledge from videos. However, their high Right-to-Wrong Rates (e.g., LongVA: 55.0\%) significantly offset these gains, indicating that they struggle to maintain their initial correct answers while processing new video information.
In contrast, human experts demonstrate effective knowledge acquisition with a higher Wrong-to-Right Rate (40.4\%) and a lower Right-to-Wrong Rate (10.7\%). This indicates humans' ability to integrate new knowledge while preserving their prior knowledge. These findings highlight a gap between human and model capabilities in video-based learning, particularly in maintaining existing knowledge while processing new information from videos.

\subsection{Error Analysis}
We analyzed the Claude-3.5-Sonnet errors in the Adaptation track by examining 100 randomly sampled error cases. Human annotators analyzed these cases to identify the root causes of mispredictions. The distribution of these errors is shown in Fig. \ref{fig:errorpiechart}, with more error cases provided in the Appendix.

\noindent \textbf{Method Selection Error (8\%)}:
The model's initial thinking direction is incorrect. For example, the model fails to adopt the correct solution strategy demonstrated in the video.

\noindent \textbf{Method Adaptation Error (64\%)}:
These represent cases where the model can correctly recall and understand the methods demonstrated in the video but fails to adapt the method to new scenarios properly. 
For example, Fig. \ref{fig:case_study} shows how models can struggle with subtle scenario differences between the video example and the Adaptation question.
While the model recalls the core DFS principles from a simple tree example in the video, it fails to adapt these principles flexibly when working with a more complex graph containing cycles. This type of error reveals its limitations in video-based learning when applying the learned methods across different contexts.

\noindent \textbf{Question Misreading Error (15\%)}:
These errors stem from misinterpreting the question requirements, such as misreading numerical values or conditions. Such errors are unrelated to the model's ability to apply knowledge from videos.

\noindent \textbf{Other Errors}: 
These include Refuse to Answer (4\%), where models express uncertainty and decline to provide an answer; Annotation error (4\%), where the annotation is inaccurate; and Answer Extraction error (5\%), where we failed to extract the answer from the longer output.

Our experiment on $\Delta_{\text{knowledge}}$ provides insights for future research in knowledge acquisition from videos:
\textbf{1)} Models showcase certain ability to acquire knowledge from videos, as indicated by their modest Wrong-to-Right Rates. However, the high Right-to-Wrong Rates often negate these gains, suggesting that models struggle to retain their initial correct reasoning when processing new information from video.
\textbf{2)} The Question Misreading and Method Selection Errors highlight the fundamental limitations in processing knowledge-intensive videos. Accurate question interpretation and a thorough understanding of video knowledge are crucial for successful knowledge application.
\textbf{3)} The significant proportion of Method Adaptation errors reveals a gap between comprehension and adaptation capabilities, suggesting that applying the knowledge from videos to solve a novel, practical scenario remains challenging for the current models.

\section{Conclusion} 
\label{sec:conclusion}

Video-MMMU systematically evaluates how large multimodal models (LMMs) acquire knowledge from videos through three cognitive stages: Perception, Comprehension, and Adaptation. Through our proposed $\Delta_{\text{knowledge}}$ metric, we reveal a gap between human and model performance, particularly in adapting acquired knowledge to novel, practical scenarios. Our insights from Video-MMMU underscore the critical need for future research to enhance LMMs' ability to learn and apply knowledge from video content effectively.

% Table \ref{tab:domainresult} presents the model performance by discipline. Overall, the closed-sourced models achieves better performance compared to open-sourced models. 

% We present the evaluation results in two tables: Table \ref{tab:domainresult} presents the model performance by discipline, and Table \ref{tab:trackresult} presents the model performance by tracks. We summarize our findings as follows:

% \subsubsection{Performance gap between human and models}
%  Human experts can achieve an overall score of 90\%, which significantly outperforms all the models reported in Table \ref{tab:domainresult}. This result highlights the existing gap between human expertise and the performance of current LMMs on Video-MMMU, emphasizing the difficulty of learning from professional-domain video lectures.

% \subsubsection{Performance across Disciplines} 
% The performance of models varies significantly across different fields. In areas such as Art, and Humanities, videos mainly introduce facts and concepts, making it easier for models to process the information. Consequently, models tend to perform better in these disciplines, demonstrating a greater capability to assimilate factual and conceptual knowledge from video content.

% In contrast, disciplines such as Science, Engineering, Business, and Medicine often involve complex reasoning, intricate calculations, dense information, and detailed visual elements like diagrams, charts, and handwritten notes. These complexities increase the difficulty of acquiring knowledge from the video lectures, resulting in lower model performance.

{
    \small
    \bibliographystyle{ieeenat_fullname}
    \bibliography{main}
}

% \clearpage
% \maketitle
% \input{sections/8.Appendix}


\clearpage
\setcounter{page}{1}
\maketitlesupplementary

\section{Subjects by Discipline}

\begin{table}[h]
\renewcommand{\arraystretch}{1.2} % Adjust row spacing
\small
\centering
\begin{tabular}{@{}>{\raggedright\arraybackslash}p{0.12\textwidth} >{\raggedright\arraybackslash}p{0.32\textwidth}@{}}
\toprule
\textbf{Discipline} & \multicolumn{1}{c}{\textbf{Subjects}} \\
\midrule
Art & Art History, Art Theory, Design, Music \\[5pt]
Business & Accounting, Economics, Finance, Manage, Marketing \\[5pt]
Science & Biology, Chemistry, Geography, Math, Physics \\[5pt]
Medicine & Basic Medical Science, Clinical Medicine, Diagnostics and Laboratory Medicine, Pharmacy, Public Health \\[5pt]
Humanities & History, Literature, Psychology, Sociology \\[5pt]
Engineering & Agriculture, Architecture and Engineering, Computer Science, Electronics, Energy and Power, Materials, Mechanical Engineering \\
\bottomrule
\end{tabular}
\caption{Subjects categorized under six disciplines.}
\label{table:subjects_two_columns}
\end{table}

\section{Additional Knowledge Acquisition Experiment Results}
We present the results of the $\Delta_{\text{knowledge}}$ experiment in Table \ref{tab:deltaresultsformoremodels}. This table includes a detailed breakdown of the number of questions that transitioned from Wrong-to-Right and Right-to-Wrong, along with the corresponding rates.

The $\Delta_{\text{knowledge}}$ metric reveals a gap between human experts and models, particularly in their ability to learn new information from videos. This skill, which humans exhibit naturally through video-based learning, arises from our long-standing reliance on videos as a medium for acquiring knowledge. Humans have developed a proficiency in extracting, retaining, and applying information from visual content, making video learning an essential component of natural knowledge acquisition.

For models to perform effectively in real-world environments, the ability to learn and adapt from videos is crucial. This capability would allow models to continuously evolve and refine their understanding, thereby enhancing their utility in dynamic and complex scenarios. 
However, the result suggests that current models are not yet capable of effectively acquiring new knowledge from video and applying it in practice. This suggests that future research needs to focus on improving how models acquire knowledge from videos - specifically, their ability to understand, remember, and apply information from video content. These improvements will be crucial for future LMMs to work effectively in the wild.

\begin{table*}[t]
\centering
\begin{tabular}{l c c c c c}
\hline
\textbf{Model} & \textbf{$\Delta_{\text{knowledge}}$ (\%)} & \multicolumn{2}{c}{\textbf{Wrong-to-Right}} & \multicolumn{2}{c}{\textbf{Right-to-Wrong}} \\
\cmidrule(lr){3-4} \cmidrule(lr){5-6}
 & & \textbf{No. of Questions} & \textbf{Rate (\%)} & \textbf{No. of Questions} & \textbf{Rate (\%)} \\
\hline
Human Expert & 33.1 & 72 & 40.4 & 13 & 10.7 \\
GPT-4o \cite{openai2024gpt4o} & 15.6 & 44 & 28.0 & 19 & 13.3 \\
Claude-3.5-Sonnet \cite{sonnet} & 11.4 & 42 & 28.8 & 30 & 19.5 \\
VILA-1.5-40B \cite{lin2023vila} & 9.4 & 57 & 25.2 & 34 & 45.9 \\
Gemini-1.5-Pro \cite{gemini2023family} & 8.7 & 49 & 29.5 & 33 & 24.6 \\
LLaVA-Video-72B \cite{zhang2024videoinstructiontuningsynthetic} & 7.1 & 40 & 22.0 & 29 & 24.6 \\
LLaVA-OneVision-72B \cite{li2024llava} & 6.6 & 37 & 20.6 & 28 & 23.3 \\
VILA-1.5-8B \cite{lin2023vila} & 5.9 & 48 & 20.2 & 35 & 56.5 \\
Aria \cite{aria} & 3.2 & 47 & 25.4 & 42 & 36.5 \\
MAmmoTH-VL-8B \cite{guo2024mammothvlelicitingmultimodalreasoning} & 1.5 & 48 & 23.9 & 45 & 45.5 \\
\hline
Gemini-1.5-Flash \cite{gemini2023family} & -3.3 & 39 & 23.5 & 42 & 31.3 \\
LLaVA-Video-7B \cite{zhang2024videoinstructiontuningsynthetic} & -5.3 & 35 & 18.5 & 47 & 42.3 \\
LLaVA-OneVision-7B \cite{li2024llava} & -5.6 & 36 & 18.2 & 43 & 42.2 \\
LongVA \cite{zhang2024longva} & -7.0 & 29 & 13.6 & 47 & 54.0 \\
InternVL2-8B \cite{chen2024far} & -8.5 & 46 & 24.3 & 61 & 55.0 \\
\hline
\end{tabular}
\caption{Additional Knowledge Acquisition Experiment Results with Delta (\%) values.}
\label{tab:deltaresultsformoremodels}
\end{table*}

\section{Prompt for Adaptation Track}
\begin{figure}
    \centering
    \includegraphics[width=0.45\textwidth]{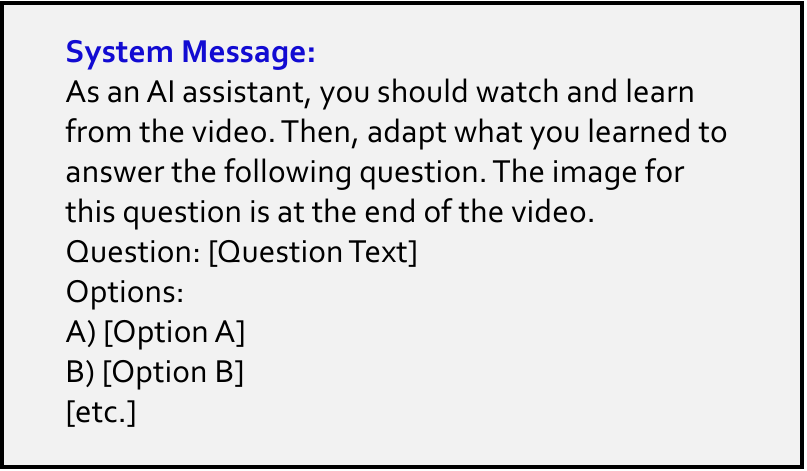}
    \caption{Prompt for Adaptation track.}
    \label{fig:prompt}
\end{figure}

In the adaptation track, we append the question's image to the end of each video. We introduce the prompt as shown in Fig. \ref{fig:prompt}.

\section{Prompt for Determining the Helpfulness of Audio}
\begin{figure*}
    \centering
    \includegraphics[width=\textwidth]{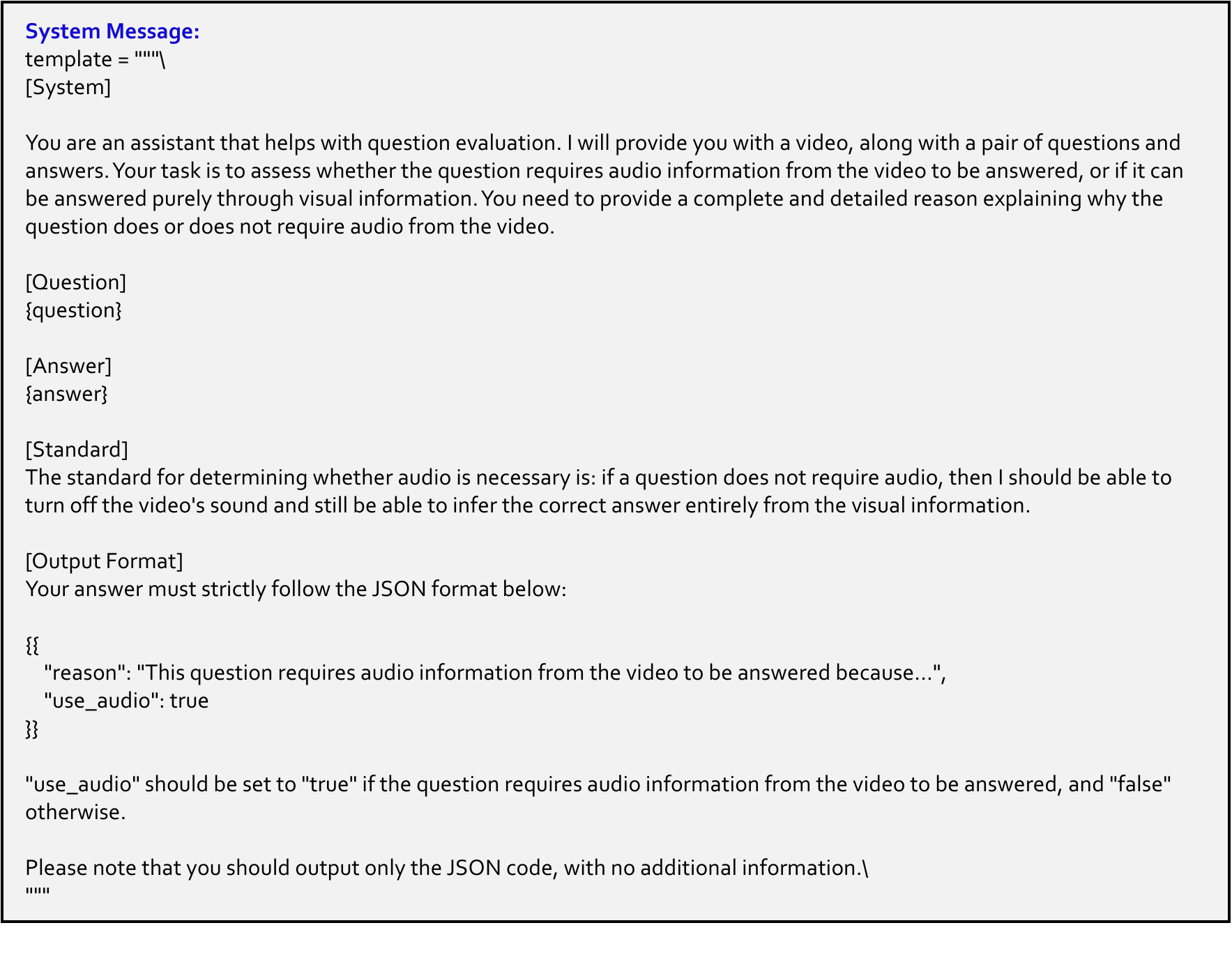}
    \caption{Prompt for determining the helpfulness of audio.}
    \label{fig:geminiprompt}
\end{figure*}

For all samples in Video-MMMU, we employ Gemini 1.5 Pro~\cite{gemini2023family} to analyze each video-question pair and determine if audio might be helpful to solve the question, as shown in Fig.~\ref{fig:figure3}\textcolor{cvprblue}{c}. This analysis will benefit more future Large Multimodal Models (LMMs) with audio processing capabilities. We introduce the prompt as shown in Fig. \ref{fig:geminiprompt}.

\section{Annotation Pipeline}
\begin{figure*}
    \centering
    \includegraphics[width=\textwidth]{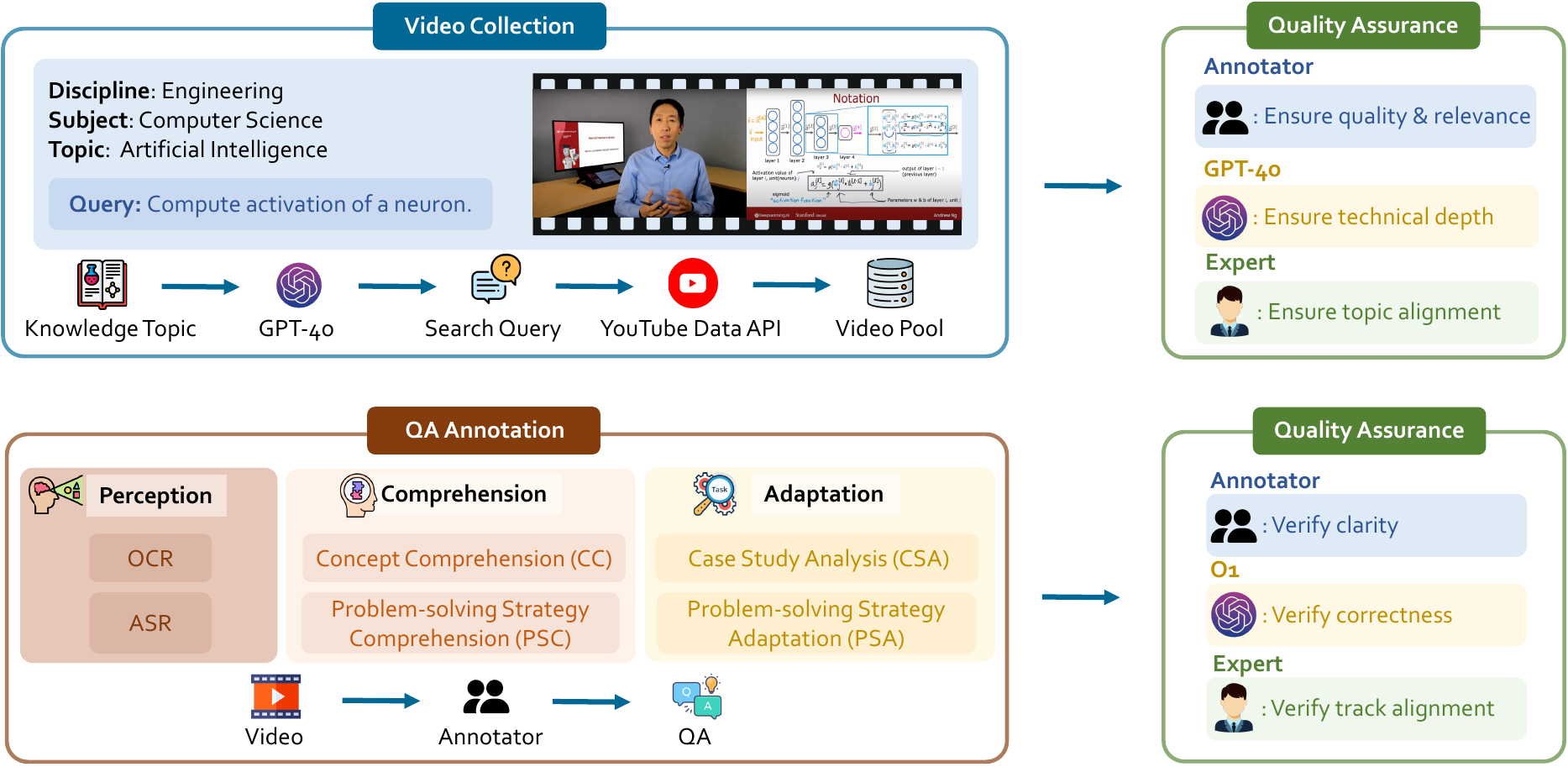}
    \caption{An illustration of the dataset curation pipeline.}
    \label{fig:pipeline}
\end{figure*}
We illustrate our pipeline for video collection and QA annotation in Fig. \ref{fig:pipeline}.

\section{More Error Analysis}
\begin{figure*}
    \centering
    \includegraphics[width=\textwidth]{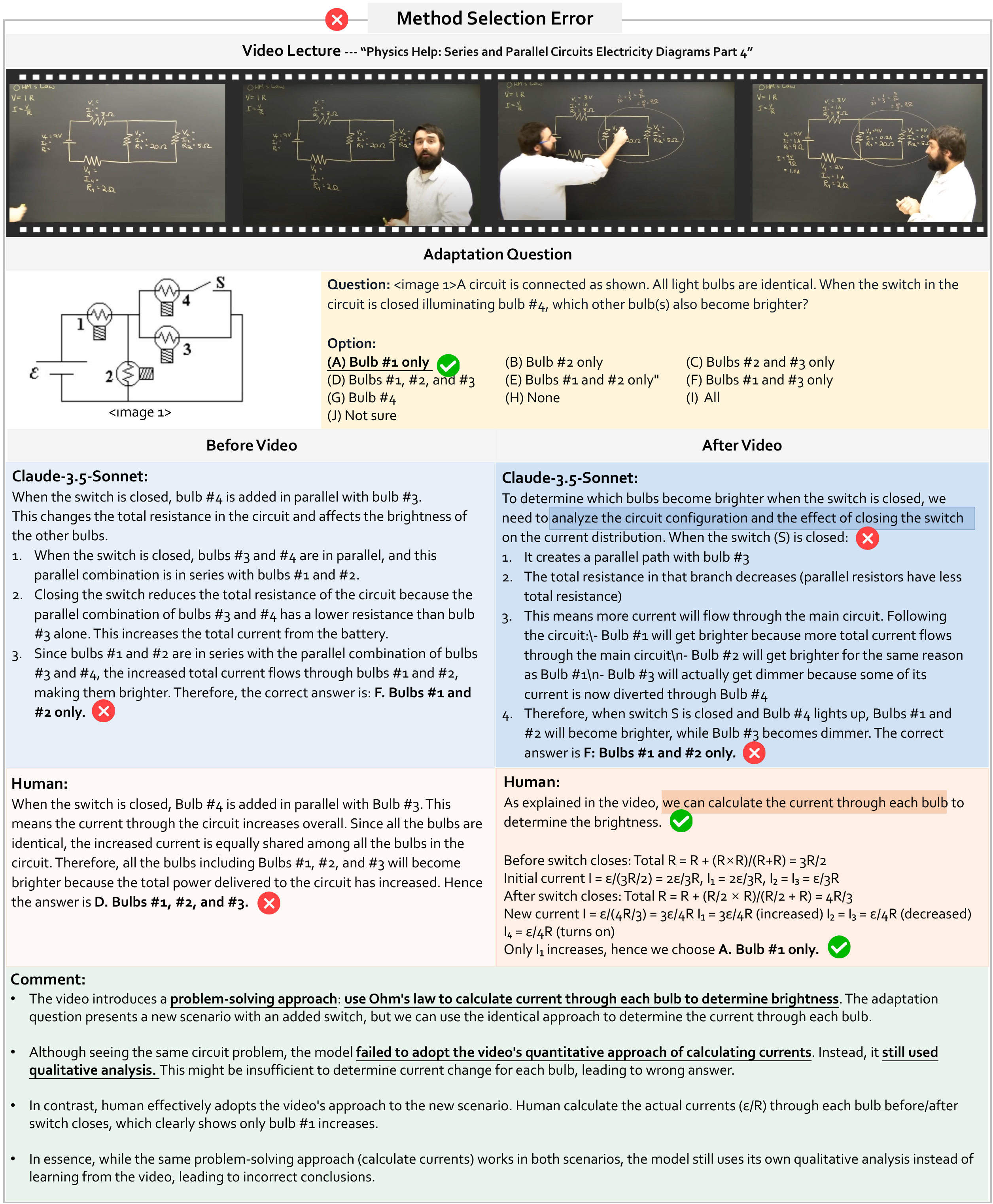}
    \caption{A sample error case in the Adaptation track: Method Selection Error by Claude-3.5-Sonnet.}
    \label{fig:methodseclectionerrorclaude_1}
\end{figure*}

\begin{figure*}
    \centering
    \includegraphics[width=\textwidth]{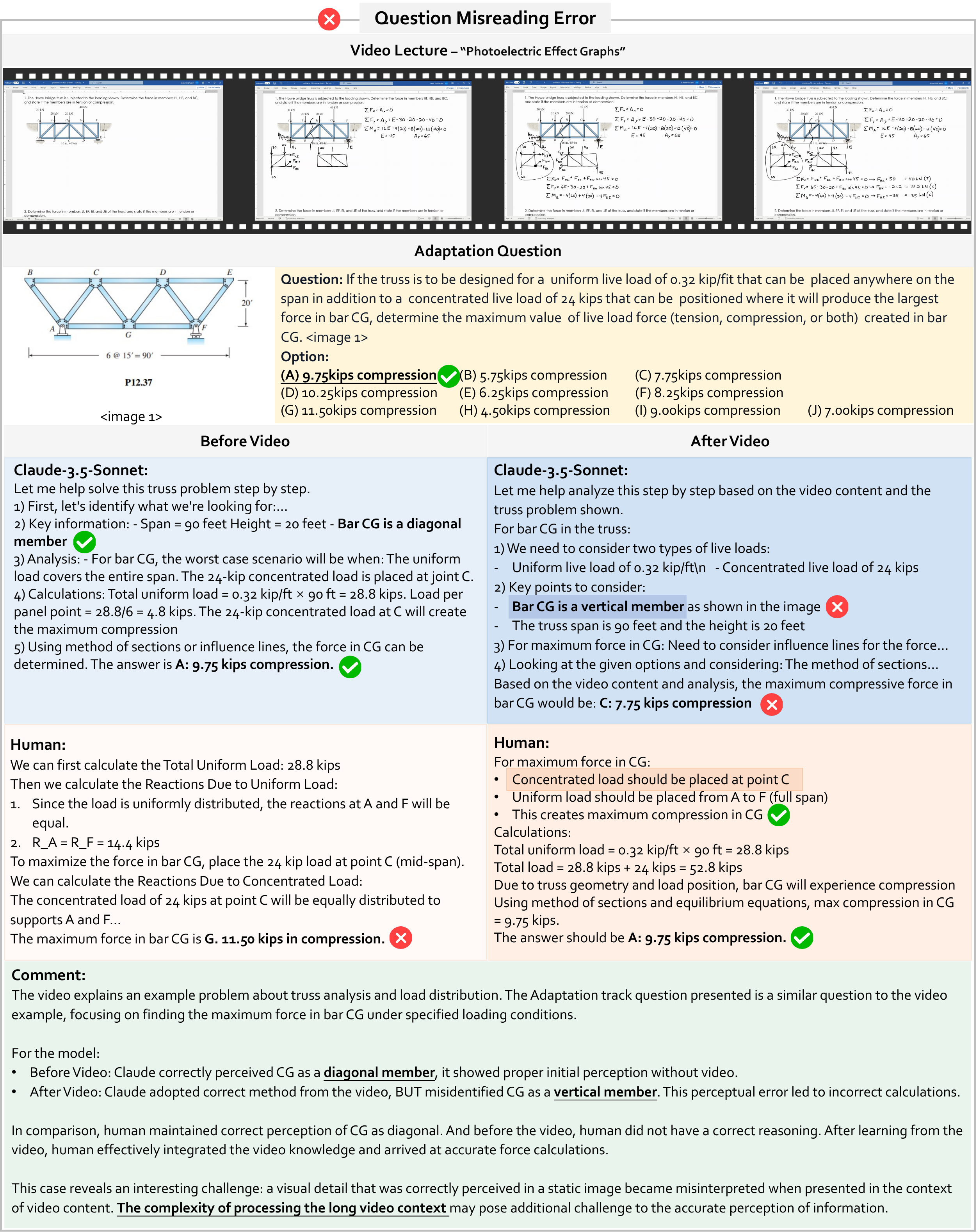}
    \caption{A sample error case in the Adaptation track: Question Misreading Error by Claude-3.5-Sonnet.}
    \label{fig:misreadingclaude1}
\end{figure*}

This section presents a comprehensive analysis of error cases across all three tracks. We begin by examining errors made by Claude-3.5-Sonnet \cite{sonnet} in the Adaptation track. Specifically, Fig. \ref{fig:methodseclectionerrorclaude_1} illustrates Method Selection Errors, while Fig. \ref{fig:misreadingclaude1} demonstrates Question Misreading Errors.

\begin{figure*}
    \centering
    \includegraphics[width=\textwidth]{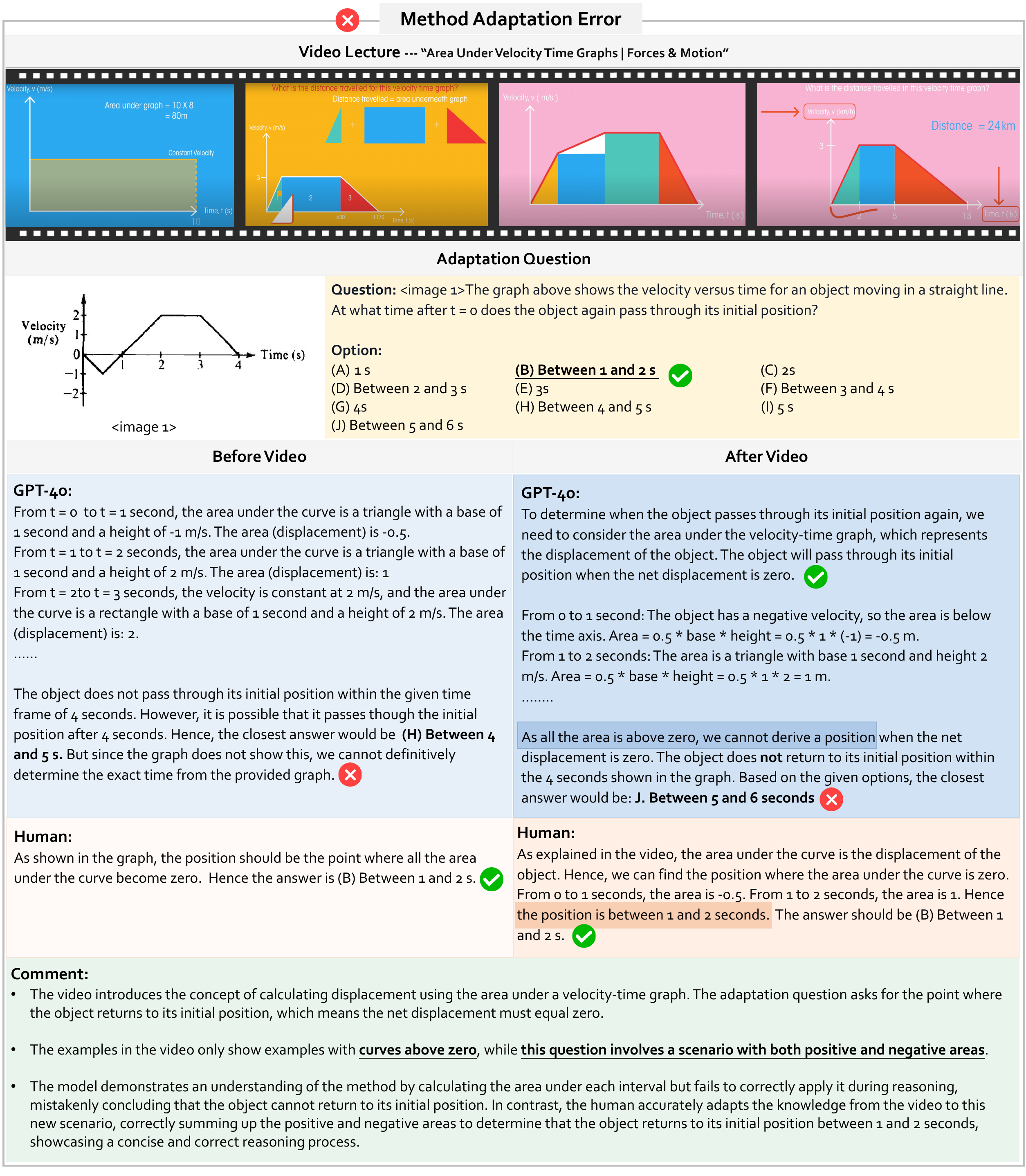}
    \caption{A sample error case in the Adaptation track: Method Adaptation Error by GPT-4o.}
    \label{fig:adaptationgpt4o_1}
\end{figure*}

\begin{figure*}
    \centering
    \includegraphics[width=\textwidth]{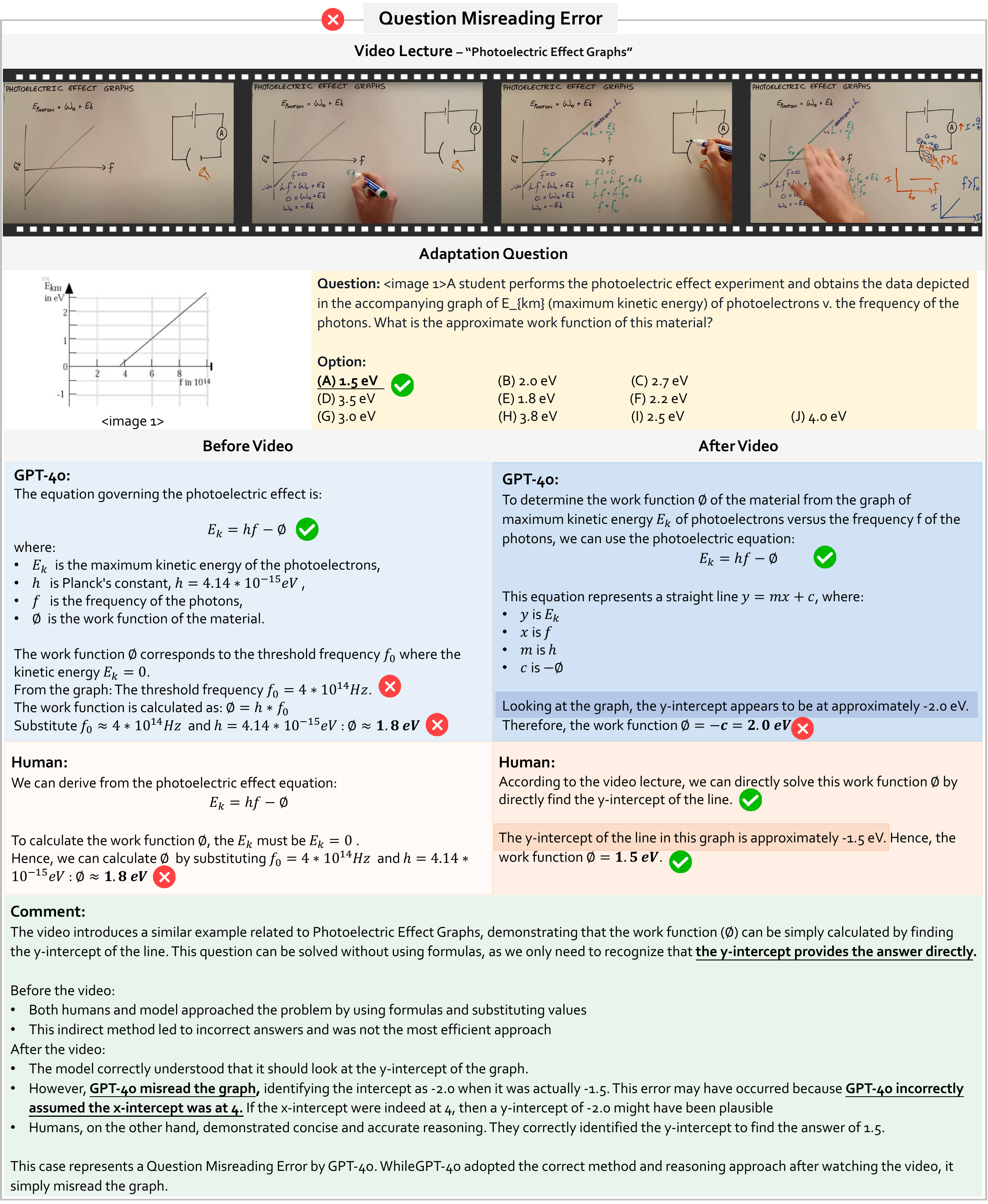}
    \caption{A sample error case in the Adaptation track: Question Misreading Error by GPT-4o.}
    \label{fig:misreadinggpt4o_1}
\end{figure*}

We also analyze error cases by GPT-4o \cite{openai2024gpt4o} in the Adaptation track. Fig. \ref{fig:adaptationgpt4o_1} and Fig. \ref{fig:misreadinggpt4o_1} present Method Adaptation Error and Question Misreading Error, respectively.

\begin{figure*}
    \centering
    \includegraphics[width=\textwidth]{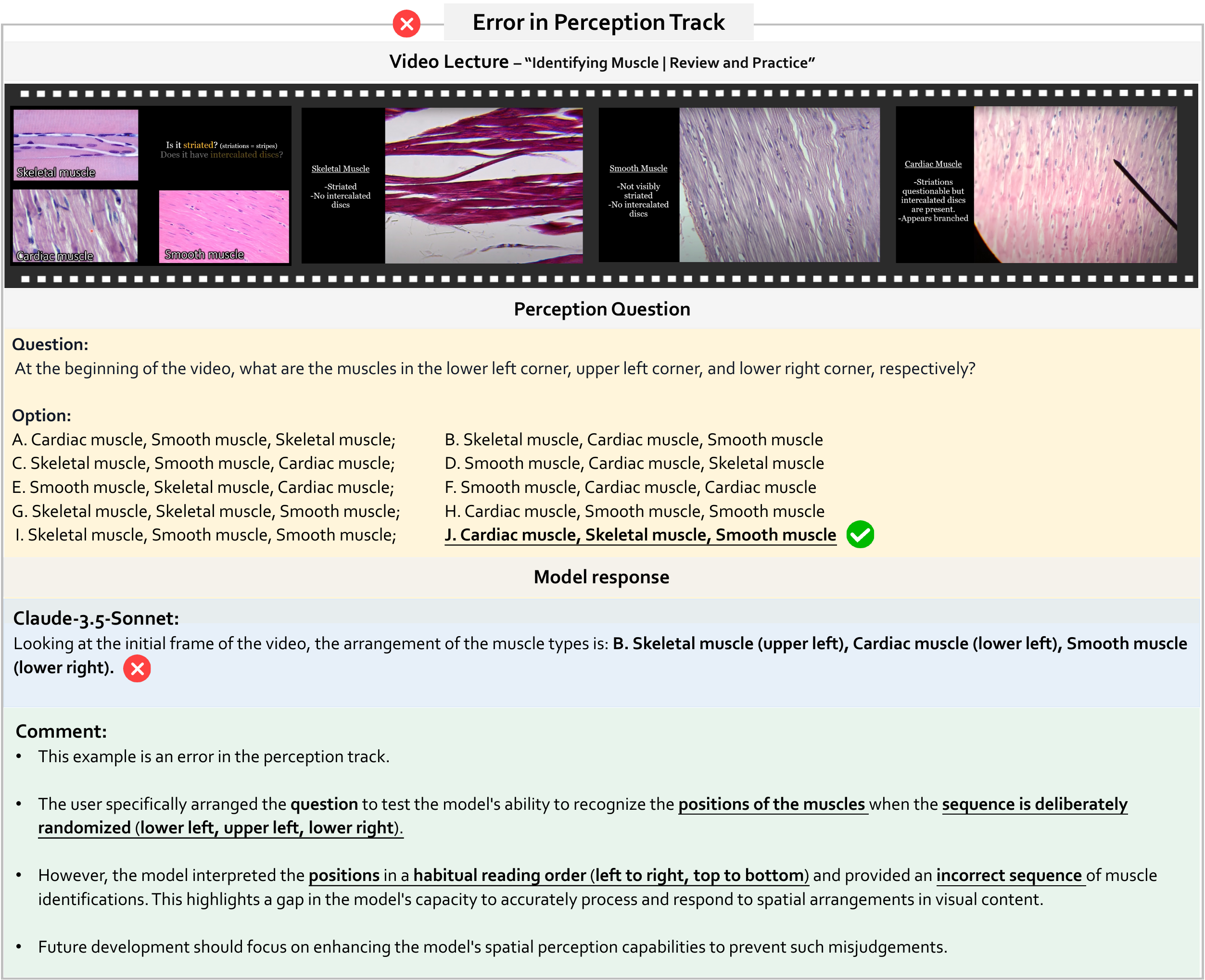}
    \caption{A sample error case in the Perception track.}
    \label{fig:perceptionclaude_1}
\end{figure*}

\begin{figure*}
    \centering
    \includegraphics[width=\textwidth]{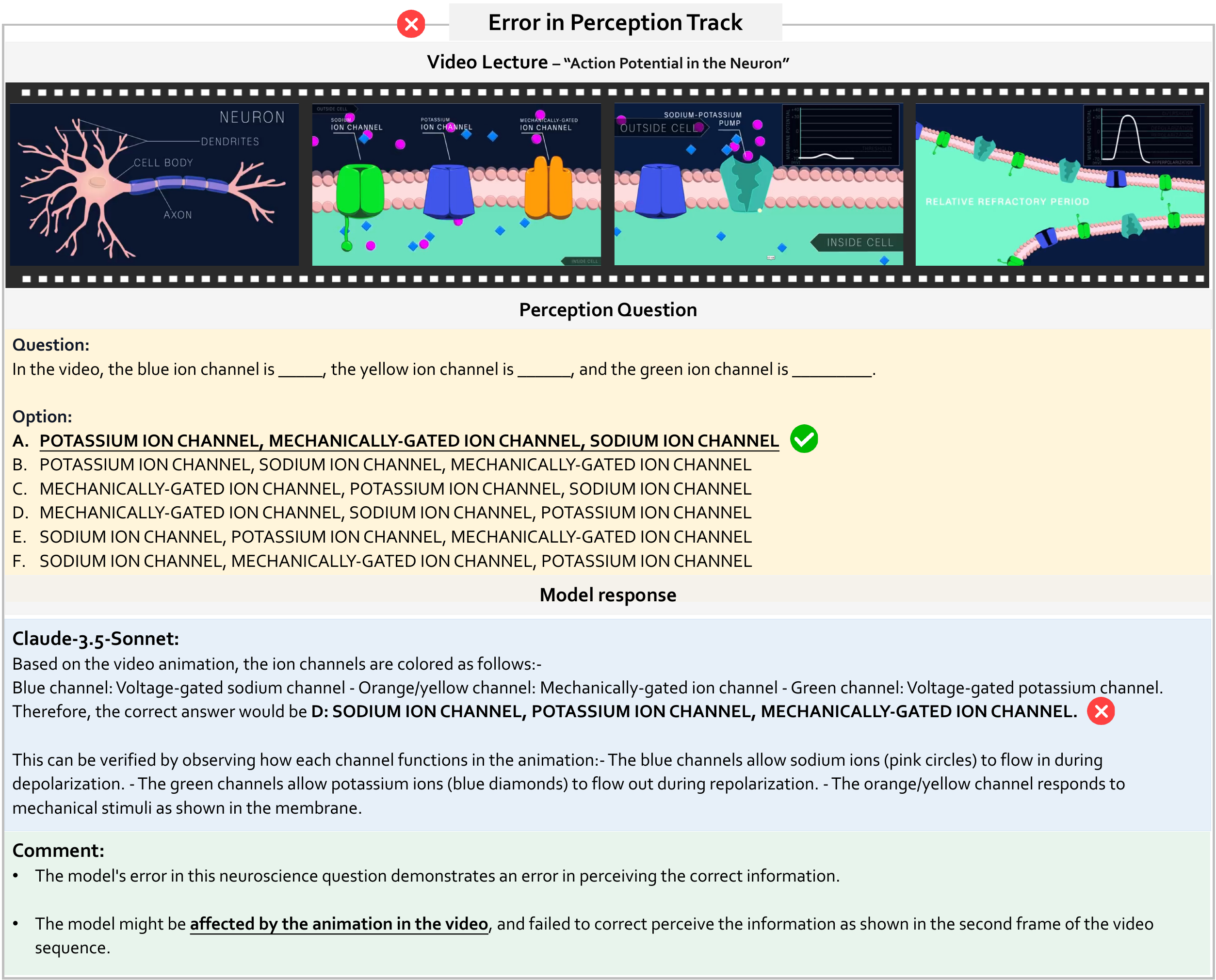}
    \caption{A sample error case in the Perception track.}
    \label{fig:perceptionclaude_2}
\end{figure*}

\begin{figure*}
    \centering
    \includegraphics[width=\textwidth]{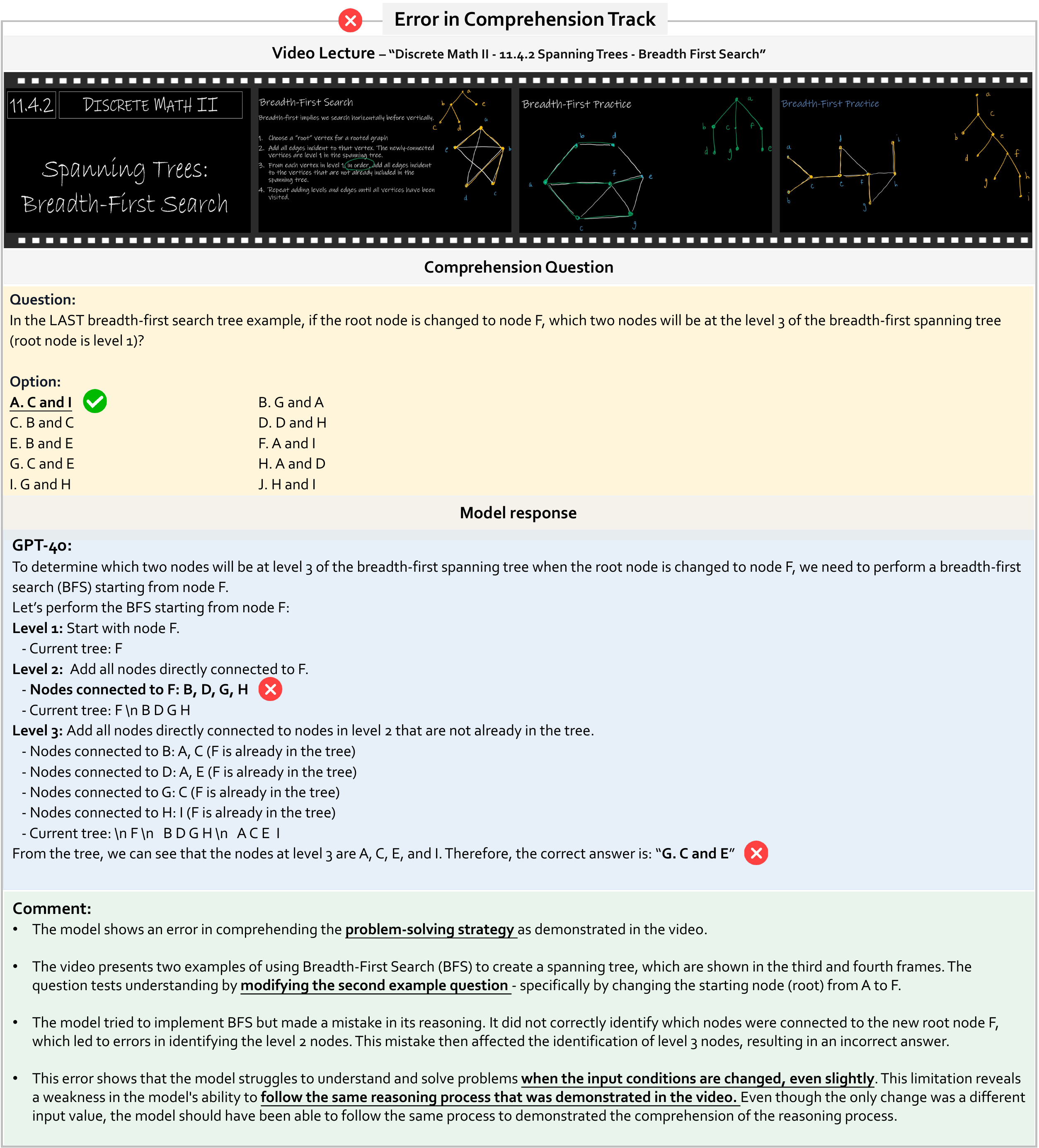}
    \caption{A sample error case in the Comprehension track.}
    \label{fig:comprehensionclaude_1}
\end{figure*}

\begin{figure*}
    \centering
    \includegraphics[width=\textwidth]{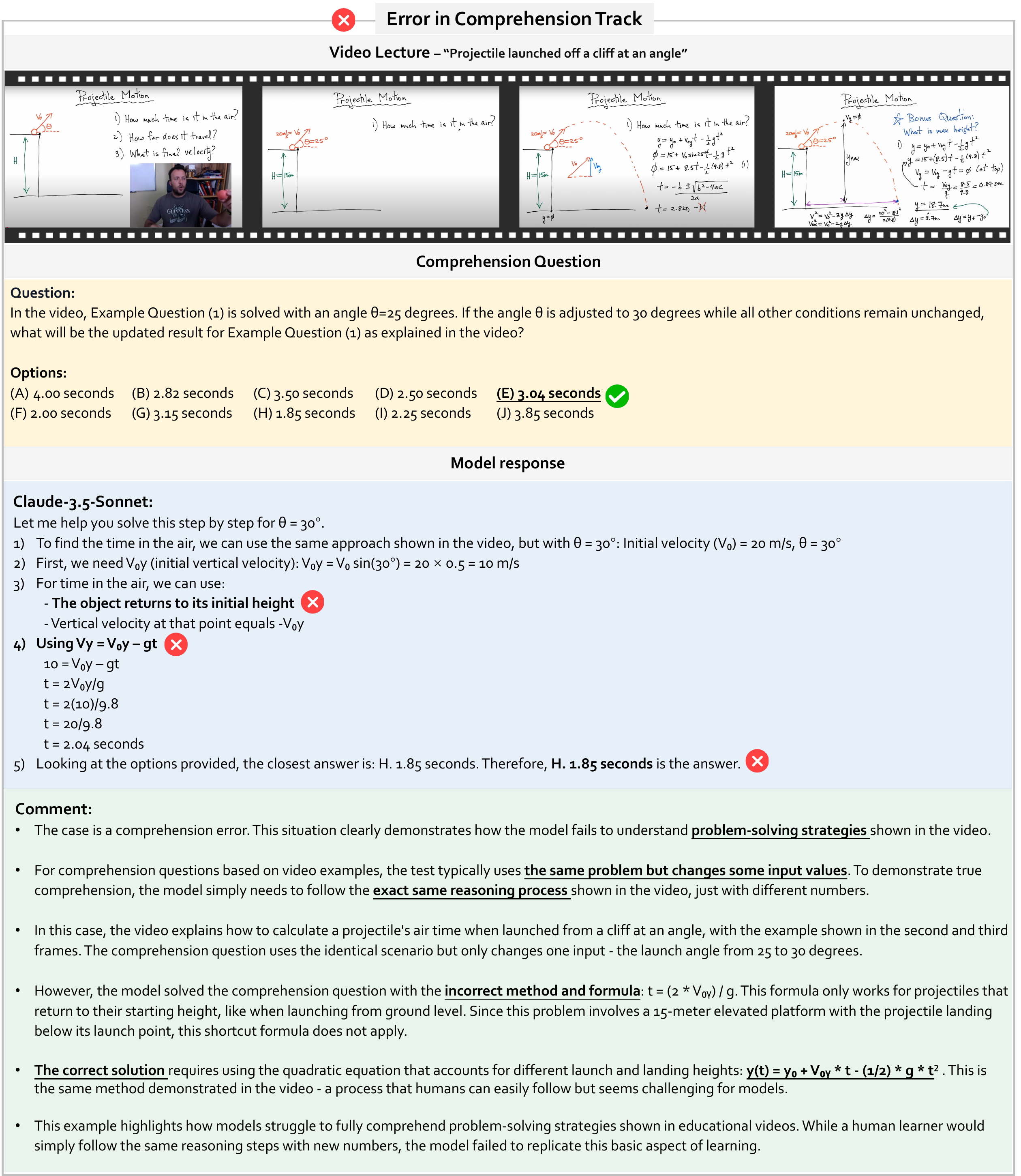}
    \caption{A sample error case in the Comprehension track.}
    \label{fig:comprehensionclaude_2}
\end{figure*}

Furthermore, we investigate error cases in both the Perception and Comprehension tracks. For the Perception track, we present two representative error cases in Fig. \ref{fig:perceptionclaude_1} and Fig. \ref{fig:perceptionclaude_2}. Similarly, for the Comprehension track, we analyze two error cases shown in FigFig. \ref{fig:comprehensionclaude_1} and Fig. \ref{fig:comprehensionclaude_2}. Each case study includes a detailed analysis of the observed errors.

\section{Wrong-to-Right Case Analysis}
\begin{figure*}
    \centering
    \includegraphics[width=\textwidth]{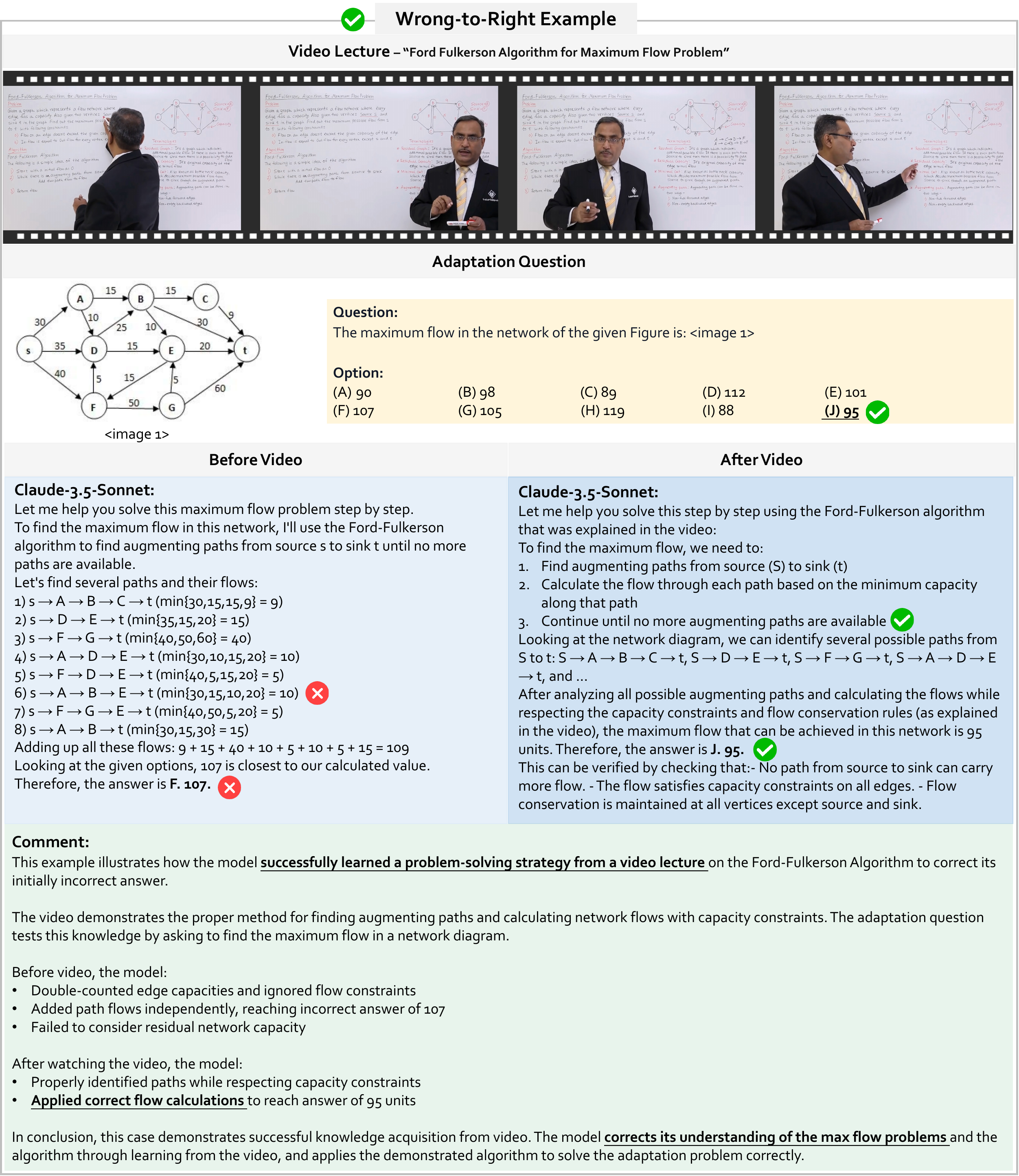}
    \caption{A Wrong-to-Right example of Claude-3.5-Sonnet in the Adaptation track.}
    \label{fig:wrongtorightclaude_1}
\end{figure*}

\begin{figure*}
    \centering
    \includegraphics[width=\textwidth]{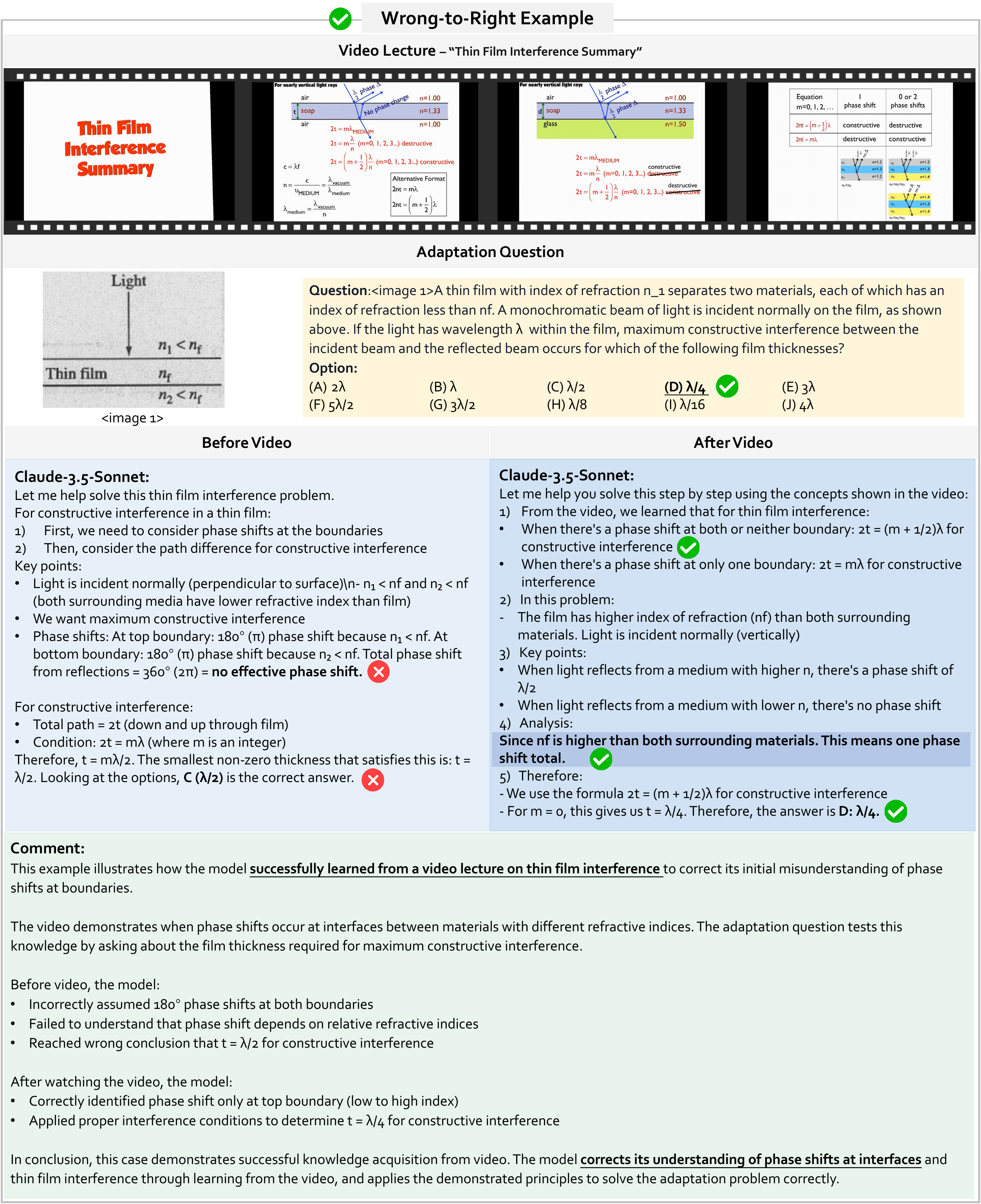}
    \caption{A Wrong-to-Right example of Claude-3.5-Sonnet in the Adaptation track.}
    \label{fig:wrongtorightclaude_2}
\end{figure*}

\begin{figure*}
    \centering
    \includegraphics[width=\textwidth]{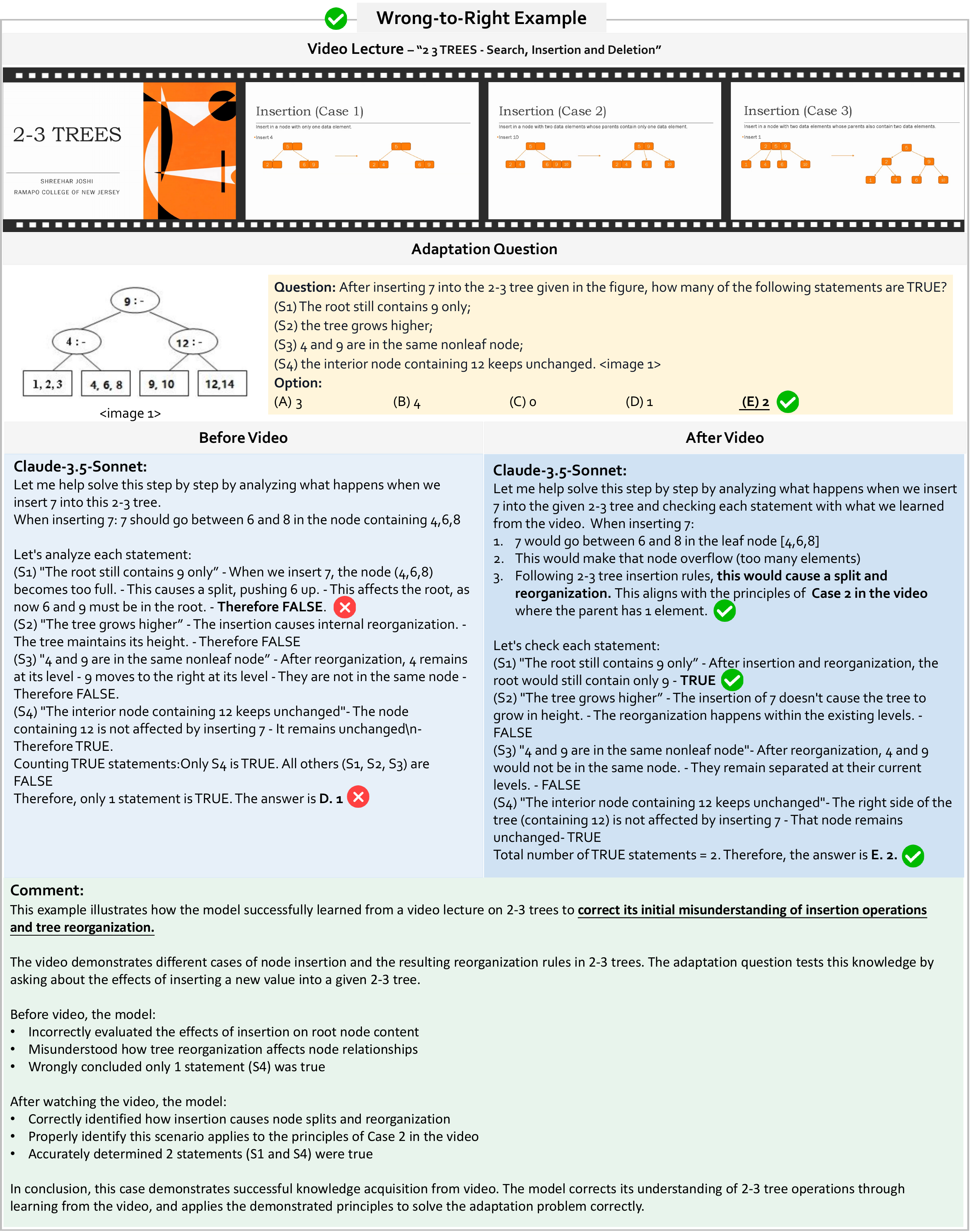}
    \caption{A Wrong-to-Right example of Claude-3.5-Sonnet in the Adaptation track.}
    \label{fig:wrongtorightclaude_3}
\end{figure*}

\begin{figure*}
    \centering
    \includegraphics[width=\textwidth]{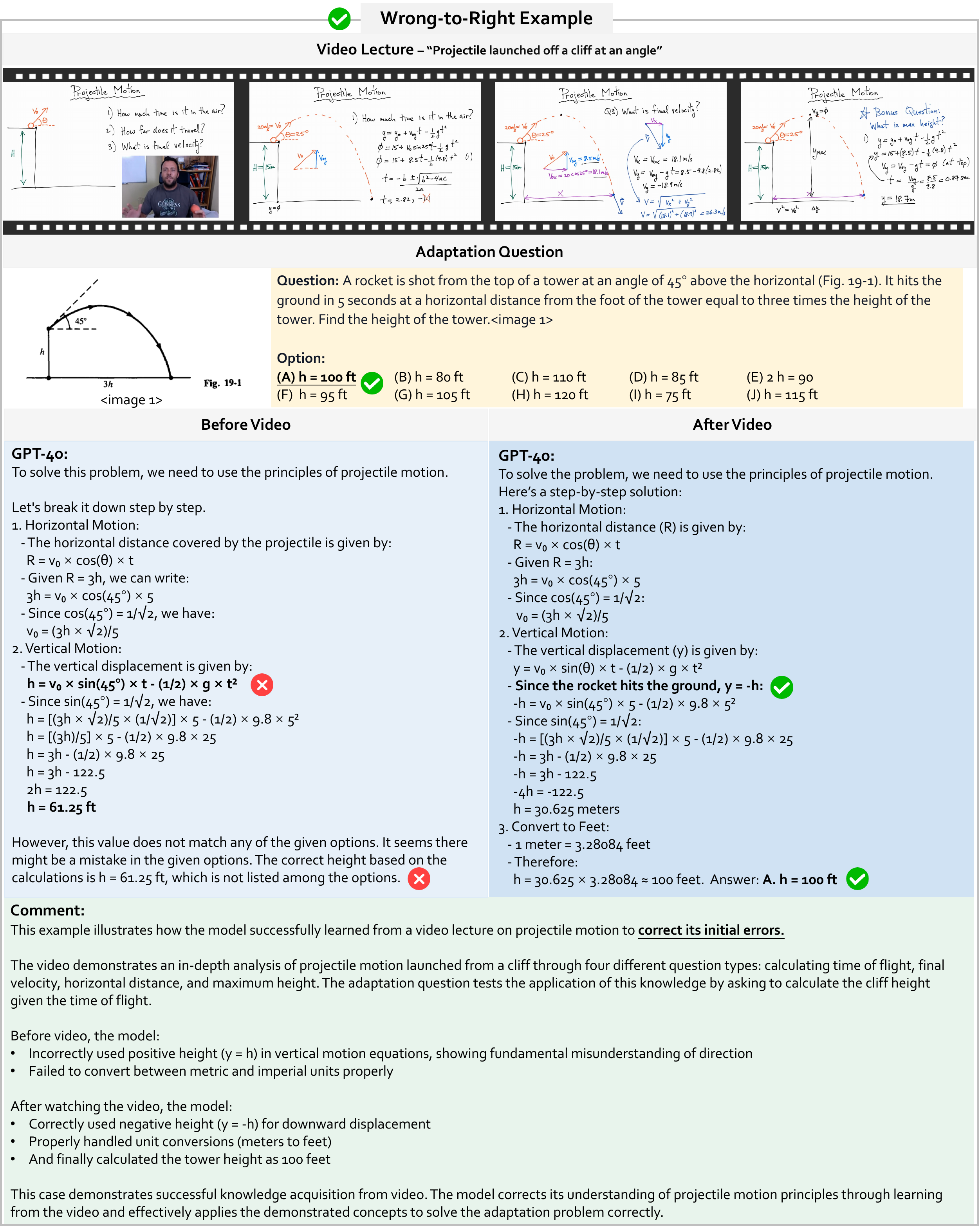}
    \caption{A Wrong-to-Right example of GPT-4o in the Adaptation track.}
    \label{fig:wrongtorightgpt4o_1}
\end{figure*}

For the Adaptation track, we also analyze the Wrong-to-Right examples where models successfully learned from video content to correctly solve Adaptation track questions. For Claude-3.5-Sonnet \cite{sonnet}, we present three such examples in Fig. \ref{fig:wrongtorightclaude_1}, Fig. \ref{fig:wrongtorightclaude_2}, and Fig. \ref{fig:wrongtorightclaude_3}. Additionally, we present a Wrong-to-Right example of GPT-4o \cite{openai2024gpt4o} in Fig. \ref{fig:wrongtorightgpt4o_1}. Each case study provides a detailed analysis of how the model successfully adapted its knowledge.

\end{document}